\documentclass{bmvc2k}

\usepackage{tikz}
\usepackage{comment}

\usepackage[accsupp]{axessibility}
\usepackage{graphicx}
\usepackage{amsmath}
\usepackage{amssymb}
\usepackage{booktabs}

\usepackage{epsfig}
\usepackage{bm}
\usepackage{multirow}
\usepackage{dsfont}
\usepackage{xcolor}
\usepackage{wrapfig}
\usepackage{color}
\usepackage{verbatim}
\usepackage{ulem}
\usepackage{enumitem}
\usepackage{placeins}
\usepackage{ragged2e}
\usepackage{diagbox}

\usepackage{textcomp}
\usepackage{soul}
\usepackage{makecell}
\usepackage{siunitx}
\usepackage{arydshln}

\useunder{\uline}{\ul}{}
\usepackage{float}

\usepackage{url}

\usepackage{hyperref}
\usepackage{breakcites}

\title{STPLS3D: A Large-Scale Synthetic and Real Aerial Photogrammetry 3D Point Cloud Dataset}

\addauthor{Meida Chen}{mechen@ict.usc.edu}{1}
\addauthor{Qingyong Hu}{qingyong.hu@cs.ox.ac.uk}{2}
\addauthor{Zifan Yu}{zifanyu@asu.edu}{3}
\addauthor{Hugues Thomas}{hugues.thomas@utoronto.ca}{4}
\addauthor{Andrew Feng}{feng@ict.usc.edu}{1}
\addauthor{Yu Hou}{yhou2@andrew.cmu.edu}{5}
\addauthor{Kyle McCullough}{mccullough@ict.usc.edu}{1}
\addauthor{Fengbo Ren}{renfengbo@asu.edu}{3}
\addauthor{Lucio Soibelman}{Soibelman@usc.edu}{6}

\addinstitution{
  University of Southern California\\
  Institute for Creative Technologies\\
  Los Angeles, CA, USA\\
}
\addinstitution{
 University of Oxford\\
 Oxfordshire, UK\\
}
\addinstitution{
  Arizona State University\\
  Tempe, AZ, USA\\
}
\addinstitution{
  University of Toronto\\
  Institute for Aerospace Studies\\
  Toronto, Ontario, Canada\\
}
\addinstitution{
  Carnegie Mellon University\\
  Pittsburgh, PA, USA\\
}
\addinstitution{
  University of Southern California\\
  Los Angeles, CA, USA\\
}
\runninghead{STPLS3D}{\STPLSHomePage}

\newcommand{\ieours}{\textit{i}.\textit{e}., }
\newcommand{\egours}{\textit{e}.\textit{g}., }
\newcommand{\nickname}{STPLS3D}
\newcommand{\qy}[1]{\textcolor{black}{#1}}

\def\dataQualityComparsionVideo{\url{https://youtu.be/4AjMWTgV2Ec}}
\def\workflowVideo{\url{https://youtu.be/6wYWVo6Cmfs}}

\def\dataDownload{\url{https://drive.google.com/file/d/17nUphM7LpmR3tT9eX0M1SXfIVjidr0-T/view?usp=sharing}}
\def\dataDownload{\url{www.stpls3d.com/data}}
\def\STPLSHomePage{\url{www.stpls3d.com}}

\begin{document}

\maketitle

\begin{abstract}
Although various 3D datasets with different functions and scales have been proposed recently, it remains challenging for individuals to complete the whole pipeline of large-scale data collection, sanitization, and annotation. Moreover, the created datasets usually face the challenge of extremely imbalanced class distribution or partial low-quality data samples. Motivated by this, we explore the procedurally synthetic 3D data generation paradigm to equip individuals with the full capability of creating large-scale annotated photogrammetry point clouds. Specifically, we introduce a synthetic aerial photogrammetry point clouds generation pipeline that takes full advantage of open geospatial data sources and off-the-shelf commercial packages. Unlike generating synthetic data in virtual games, where the simulated data usually have limited gaming environments created by artists, the proposed pipeline simulates the reconstruction process of the real environment by following the same UAV flight pattern on different synthetic terrain shapes and building densities, which ensure similar quality, noise pattern, and diversity with real data. In addition, the precise semantic and instance annotations can be generated fully automatically, avoiding the expensive and time-consuming manual annotation. Based on the proposed pipeline, we present a richly-annotated synthetic 3D aerial photogrammetry point cloud dataset, termed \href{www.stpls3d.com}{STPLS3D}, with more than 16 $km^2$ of landscapes and up to 18 fine-grained semantic categories. For verification purposes, we also provide datasets collected from four areas in the real environment. Extensive experiments conducted on our datasets demonstrate the effectiveness and quality of the proposed synthetic dataset.
\end{abstract}

\section{Introduction}
\label{sec:intro}
Small Unmanned Aerial Vehicle (sUAV) and photogrammetry technologies have witnessed dramatic development over the past few years, enabling rapid reconstruction of large terrain with several square kilometers. Compared with the airborne LiDAR mapping \cite{18,19,20}, aerial photogrammetry offers an affordable solution for 3D mapping, hence attracting widespread attention from both researchers and industry practitioners for various applications \cite{hou2021approach,chen2020semantic,32,shi2021spatial}. Recently, a handful of works \cite{23,46,57,58,54,94,hu2021learning,hu2021sqn,56,chen2020photogrammetric,chen2020fully} have started to explore the semantic understanding of large-scale 3D point clouds, with promising results and insightful conclusions achieved.

Although a number of 3D datasets \cite{1,3,10,11,13,14,15,16,17,18,19,20,21,22,23,24,25,26,hu2022sensaturban} have been proposed in the last decades, it remains highly challenging for individuals to complete the whole pipeline of the customized dataset production independently for three reasons. 1) The annotation of large-scale 3D data is labor-intensive and time-consuming. In contrast to 2D data annotation, annotating 3D data such as point clouds requires extensive training to navigate and operate in the 3D environment \cite{3}. 2) Due to the limitations of hardware configurations (\egours availability of gimbal) and survey constraints (\egours flight altitudes and overlaps between images), the reconstructed point clouds are usually relatively small in size or have low-quality data samples (non-uniform density, holes, outliers, \textit{etc.}), which may have a negative impact on the execution of subsequent tasks. 3) Considering the long-tail distribution of objects in the real world, the created datasets are likely to suffer from extremely imbalanced class distribution, which poses extra challenges for downstream tasks such as semantic understanding \cite{23}.

Motivated by this, we develop a fully automatic pipeline for controllable, high-quality, and photorealistic synthetic aerial photogrammetry 3D data generation. In particular, the rich annotations, including semantic and instance labels, can be generated effortlessly as byproducts of our pipeline. Specifically, the proposed data generation pipeline has the following appealing advantages: 1) Unlike other virtual gaming engine-based generation approaches \cite{1,63}, where only limited gaming environments created by artists are used, our pipeline fully exploits existing open geospatial data sources to set up the 3D environment, with a large variety of authentic terrain shapes and building densities. 2) Considering the homogeneous architectural styles and construction materials in real-world environments, we leverage procedural modeling tools to create building models with variations and adapted different material databases to enrich the diversity of building appearances. 3) We explicitly balance the class distribution in the real world by heuristically placing 3D models of underrepresented objects in virtual environments. 4) Lastly, instead of random points sampling or ray casting \cite{63,64,72} on the 3D surfaces, we simulate similar UAV paths over the virtual terrain as the real-world survey, followed by the photogrammetry steps to reconstruct the 3D point clouds. This ensures that the generated 3D point clouds from our pipeline have similar quality and even comparable noise as real-world aerial photogrammetry data since the exact same data collection and reconstruction processes are executed.

\begin{figure}[t]
\centering
\includegraphics[width=0.9\linewidth]{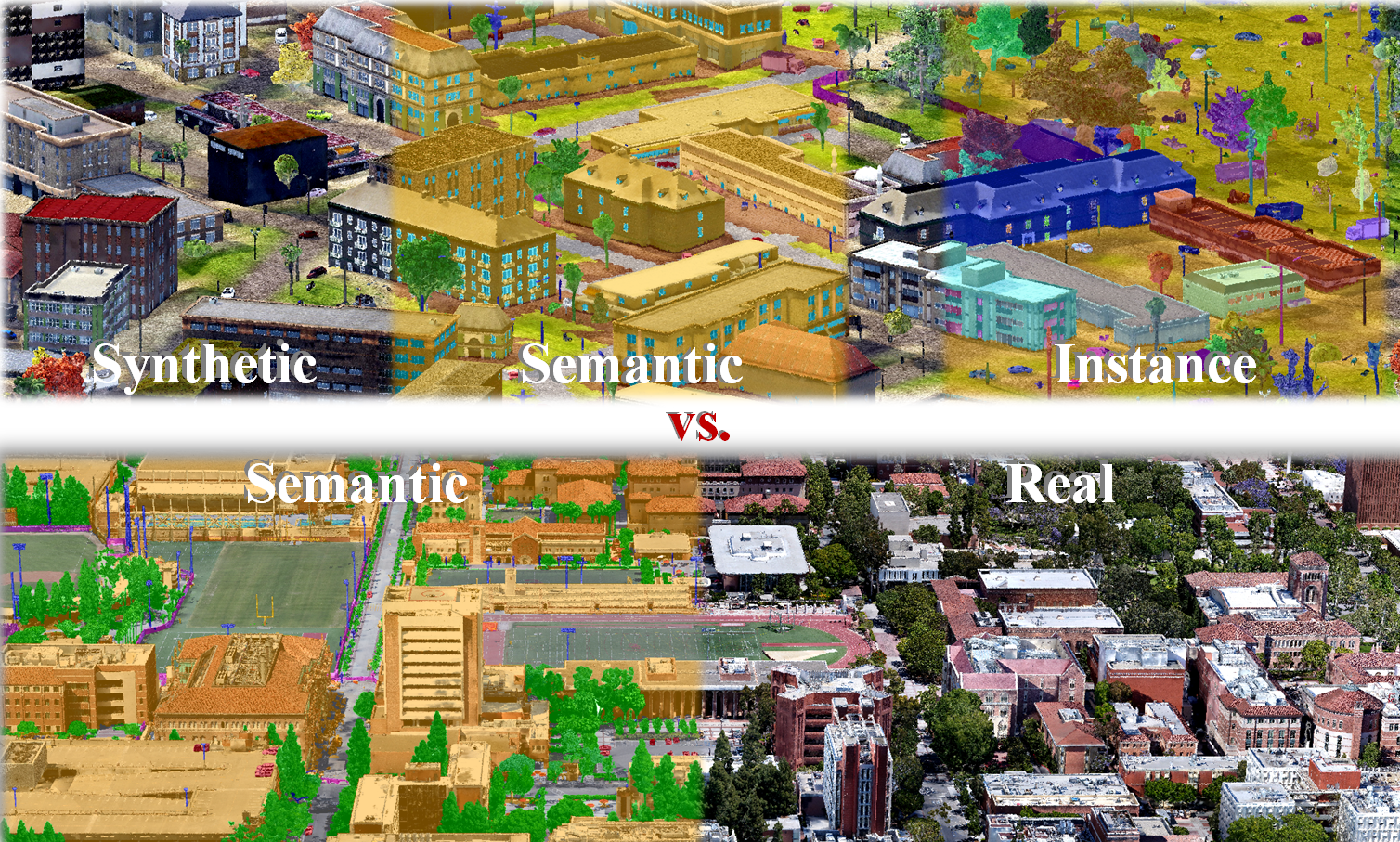}
\vspace{-0.2cm}
\caption{Example point clouds in \nickname{} dataset. Top row: synthetic point clouds with point-wise semantic and instance annotations. Bottom row: real point clouds captured from USC.}
\vspace{-0.5cm}
\label{fig: teaser}
\end{figure}

With the proposed synthetic data generation pipeline, we have further built a large-scale photogrammetry 3D point cloud dataset, termed Semantic Terrain Points Labeling - Synthetic 3D (\href{www.stpls3d.com}{STPLS3D}), which is composed of high-quality, rich-annotated point clouds from real and synthetic environments, as shown in Figure \ref{fig: teaser}. Specifically, we first collect real-world aerial images using photogrammetry best practices with quadcopter drone flight at a low altitude with significant overlaps between adjacent photos. We then reconstructed point clouds with a 1.27 $km^2$ landscape following the standard photogrammetry pipeline. Next, we follow the same UAV path and flying pattern to generate 62 synthetic point clouds with different architectural styles, vegetation types, and terrain shapes. The synthetic dataset covers about 16 $km^2$ of the city landscape, with up to 18 fine-grained semantic classes and 14 instance classes. Extensive experiments were conducted on our \nickname{} dataset to validate the quality and function of the synthetic dataset. In particular, by incorporating our synthetic dataset into the training pipeline, existing deep neural architectures can achieve visible improvement on the real data, even without adopting any domain adaptation techniques. To summarize, the main contributions of our paper are listed as follows:

\vspace{-0.1cm}
\begin{itemize}
\setlength{\itemsep}{0pt}
\setlength{\parsep}{0pt}
\setlength{\parskip}{0pt}
    \item We built a unique, richly-annotated large-scale photogrammetry point clouds dataset with synthetic and real subsets, covering more than 17 $km^2$ of the city landscape.
    \item We introduce a fully automatic pipeline for controllable, high-quality, and photorealistic synthetic aerial photogrammetry 3D data generation.
    \item Extensive experiments demonstrate the quality and function of the generated synthetic data.
\end{itemize}
\vspace{-0.3cm}

\section{Related works}
\label{sec:related_work}
Here, we provide a brief overview of existing 3D datasets; for comprehensive surveys, please refer to \cite{2,19,20,23,24,guo2020deep}. \noindent\textbf{3D Real-World Datasets.} Thanks to the development of remote sensing technologies, considerable efforts have been devoted to building 3D datasets and benchmarks for semantic understanding. To capture 3D rich geometry of the real environments, previous works usually adopted RGB-D sensors \cite{6,7,8,9,75,76} for indoor 3D scenes and utilized terrestrial scanners \cite{3,11,77}, mobile scanners \cite{10,12,13,14,15,16,22,79,80,81,82,83}, and aerial laser scanners\cite{17,18,19,20,KOLLE2021100001} for outdoor environments. Additionally, researchers from remote sensing communities also collected large-scale 3D scene-based datasets (\egours construction sites) \cite{30,31} through photogrammetry techniques with quadcopter drones and fixed-wing UAVs as the main platform. In particular, a handful of recent works have started to mount multiple sensors together on UAVs for efficient data collection in a large district \cite{32,33,34}. Overall, the scale of recent datasets has become increasingly large, and the content covers sufficient information for multiple purposes. However, due to the survey configuration and the specific photogrammetry software \cite{23,24} used, noticeable drawbacks could be found in the existing released datasets, such as missing points on the vertical surfaces, large holes, and non-uniform point density, \textit{etc.} In addition, insufficient and incorrect annotations are another common issue that could deteriorate the quality of the dataset, further leading to the inability to fairly and comprehensively evaluate the performance of deep neural models for subsequent tasks.

\begin{table}
\begin{center}
\caption{Comparison with the representative aerial datasets used for segmentation of 3D point clouds. \protect\textsuperscript{1}The number of categories with instance labels, \protect\textsuperscript{2}Labeled area.}
\vspace{-0.2cm}
\setlength{\tabcolsep}{5pt}
\resizebox{1.0\textwidth}{!}{%
\begin{tabular}{rcccccc}
 \Xhline{2.0\arrayrulewidth}
Name and Reference & \# Semantic & \# Instance\textsuperscript{1}  & \# Views / scenes & 2D Annotations & Area\textsuperscript{2} ($km^2$) & Sensor  \\ \hline
DublinCity \cite{18}& 13  & No & 8,504 / 2 & No & 2& \multirow{3}{*}{Aerial LiDAR}   \\
DALES \cite{19} & 8 & No & 1 large scene & -  & 10 & \\
LASDU \cite{20} & 5 & No & 1 scene & -  & 1.02 & \\
Swiss3DCities \cite{25} & 5 & No & 3 scenes  & No & 2.7  & quadcopter + photogrammetry \\
Campus3D \cite{24}  & 14  & 4 classes  & 6 scenes  & No & 1.58 & quadcopter + photogrammetry \\
SensatUrban \cite{23} & 13  & No & 3 scenes  & No & 4.4 & fixed wing + photogrammetry \\ \cdashline{1-7}
\textbf{STPLS3D - Real} & 6 & No & 16,376 / 4  & Yes& 1.27 & quadcopter + photogrammetry \\
\textbf{STPLS3D - SyntheticV1} & 5 & No & \textbf{17,164} / 14 & Yes& 4.22 & Synthetic Aerial photogrammetry \\
\textbf{STPLS3D - SyntheticV2} & 17  & 14 classes & 13,229 / 24 & Yes& 5.76 & Synthetic Aerial photogrammetry \\
\textbf{STPLS3D - SyntheticV3} & \textbf{18}  & \textbf{14 classes} & 15,888 / \textbf{25} & Yes& \textbf{6} & Synthetic Aerial photogrammetry \\
\Xhline{2.0\arrayrulewidth}
\end{tabular}%
}
\vspace{-0.7cm}
\label{tab:Overview-dataset-1}
\end{center}
\end{table}

\noindent\textbf{3D Synthetic Datasets.} Due to the expensive data collection and annotation costs, several works have explored the possibility of creating replaceable 3D synthetic data. Specifically, earlier works typically focused on creating synthetic point clouds for individual objects \cite{4,5,65,73,74}, while recent works have started to investigate the synthetic generation of the outdoor 3D point clouds in virtual gaming environments \cite{1,26,63,64,70,71,72,chen2020generating,SynLiDAR,le21wacv}. However, the geometrical structure, noise pattern, and sampling scheme of these datasets are still different from the real environment, leading to visible domain gaps. Additionally, since the gaming environments were manually created by artists and designers, the spatial scale of existing synthetic datasets is also limited. By contrast, we explore the outdoor large-scale 3D scene synthesis from aerial views and photogrammetry techniques with procedurally generated virtual environments. Table \ref{tab:Overview-dataset-1} compares the statistics of the proposed \nickname{} with a number of existing aerial 3D datasets.

\section{Synthetic Data Generation Pipeline}
\label{section:3}
\begin{figure}[t]
\centering
\includegraphics[width=1.0\linewidth]{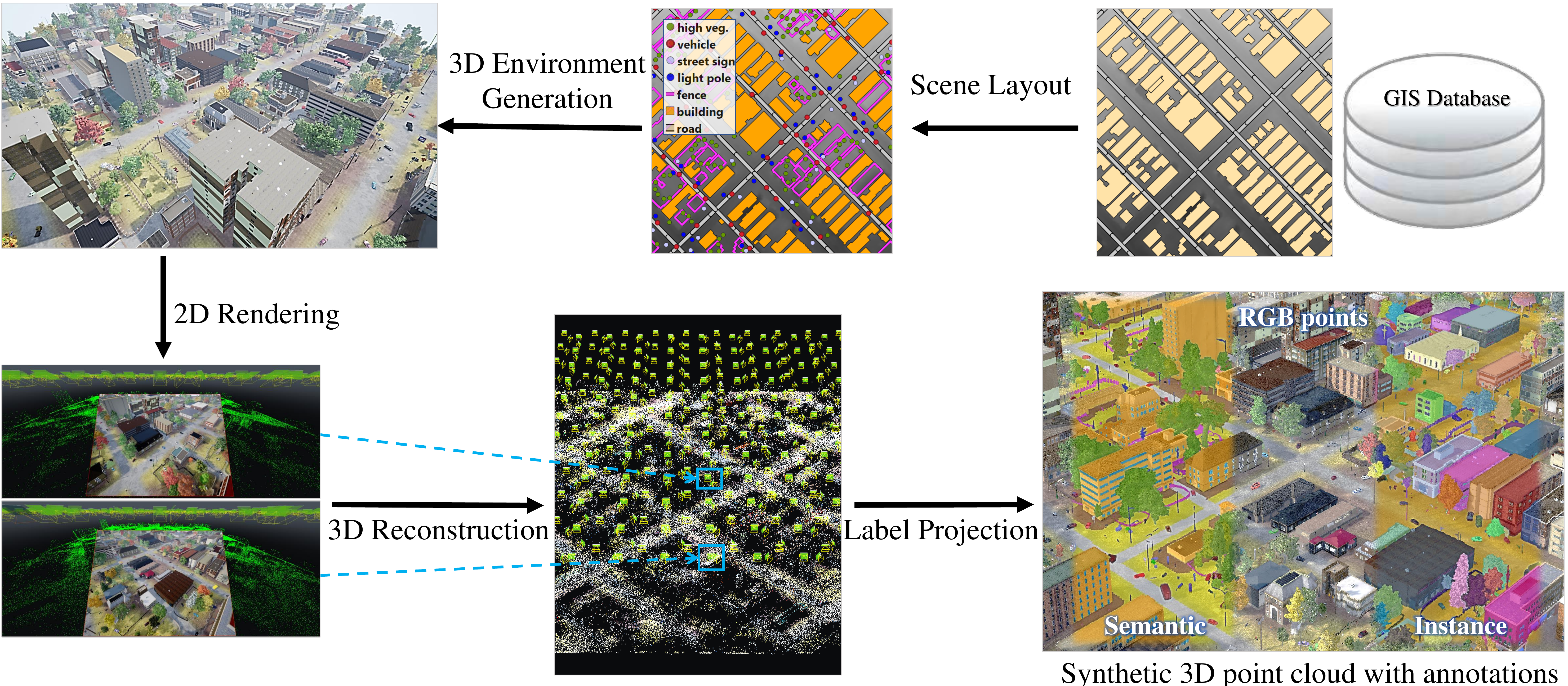}
 \vspace{-0.3cm}
\caption{The proposed synthetic data generation pipeline.}
 \vspace{-0.2cm}
\label{fig: pipeline}
\end{figure}

The synthetic data generation pipeline is illustrated in Figure \ref{fig: pipeline}. Overall, the main idea is to replicate the steps one would take when creating aerial photogrammetry point clouds in the real world. In particular, we focused on bringing 3D virtual assets in the simulation that is close to reality and reconstructing point clouds with similar quality as the real ones to minimize the domain gap between synthetic and real data as much as possible. 

\subsection{Procedural 3D Environment Generation}
\label{sec3.1}

To ensure the placements of objects in the virtual environments roughly follow the form of a real city block, we built 3D virtual environments based on the Geographic Information System (GIS) data sources \cite{68} (\ieours building footprints, road networks, and digital surface models) that are publicly available. Specifically, 3D road segments are placed and extended along the road vectors obtained from the Open Street Map (OSM) \cite{69}. 3D trees, vehicles, and other city furniture are placed in the scene using predefined strategies to increase its realism and diversity. Distance constraints are also heuristically introduced during object placements to avoid unrealistic situations such as the intersection between objects. Additionally, such constraints are also used to ensure the locations of each object are contextually reasonable, \ieours the vehicles, street signs, and light poles will be on or near the roads. Besides, placing the trees pure randomly throughout the environment may produce unnatural results. Therefore, trees are placed in clusters within polygonal areas  procedurally generated as boundaries to simulate forests. In addition, individual trees are also placed around the buildings within a buffer to simulate the residential blocks. Please refer to Appendix \ref{Object placement principles for creating the synthetic environment layouts} for more details on our designed object placement principles.

This study used Computer Generated Architectural (CGA) shape grammar of CityEngine-based tools to create 3D building models based on the OSM building footprints. The procedural tool automatically extruded the footprints and added architectural elements. The overall façades generation and architectural element placements allow various types of 3D buildings to be generated from the same building footprint with different predefined CGA rules. Both the building types and heights were randomly assigned during the building generation process to ensure the synthetic environments cover a large spectrum of building variations.

\subsection{2D Image Rendering and 3D Reconstructions}
\label{2D Image Rendering and 3D Reconstructions}

The naive solution to generating a point cloud with the created 3D environments would be either directly sampling points on the 3D model surface or using a ray casting approach with predefined camera parameters. However, it produces point clouds that perfectly match the 3D virtual environment, which does not have the same quality and noise level as the data that was collected from the real world. To reduce the domain gap that exists between the sampled or ray-casted points and the real-world aerial photogrammetry point clouds, we propose to first render the 2D images in Unreal Engine 4 (UE4) using the AirSim simulator \cite{59}. In particular, we utilized weather effects to simulate fog, wind and changing sunlight directions, so as to generate more realistic 2D images from the virtual environment. With the rendered 2D images, we then reconstructed the 3D point clouds using the off-the-shelf commercial photogrammetry software (\textit{i.e.}, Bentley ContextCapture). In particular, we keep the software consistent with that of reconstructing point clouds from real-world photos. Please refer to Appendix \ref{sec:ComparisonofDataQuality} for the intuitive quality comparison between the ray-casting 3D points, the synthetic photogrammetric points, and the real-world photogrammetric point clouds.

\subsection{Semantic and Instance Annotation}
Finally, the generated synthetic point clouds are enriched with semantic and instance annotations that are automatically generated while rendering the 2D images. Note that, due to the noises introduced from the photogrammetry reconstruction process, directly casting the 2D labels to the photogrammetry point cloud will create misaligned annotations. To this end, we transfer the rendered 2D annotations to the photogrammetry point clouds with the following two steps. First, we create a proxy 3D point cloud using the ray casting method with the known intrinsic and extrinsic camera parameters and depth maps. Next, we transfer the labels from the proxy 3D point cloud to the photogrammetry points through a nearest-neighbor search algorithm, with the constraint that ground points are connected to form a large connected component and reduce the inconsistent projections due to the simulated wind effects. Though a small amount of mislabeled points may still occur at the boundaries between different objects, they did not have a significant impact while training the segmentation models in our experiments.

As shown in Figure \ref{fig: pipeline}, the proposed pipeline can generate synthetic point clouds with semantic and instance annotations. It is worth mentioning that the instance annotations are very useful for tasks such as vegetation identification (\egours tree segmentation \cite{erikson2003segmentation}) and forest management (\egours automatic tree counting \cite{li2017deep}) since it is highly challenging or even infeasible to obtain precise instance labels in the real data (\egours hundreds of thousands of overlapped trees need to be manually segmented from forest areas).

\section{Datasets}

\begin{figure*}
\centering
	\includegraphics[width=1.0\linewidth]{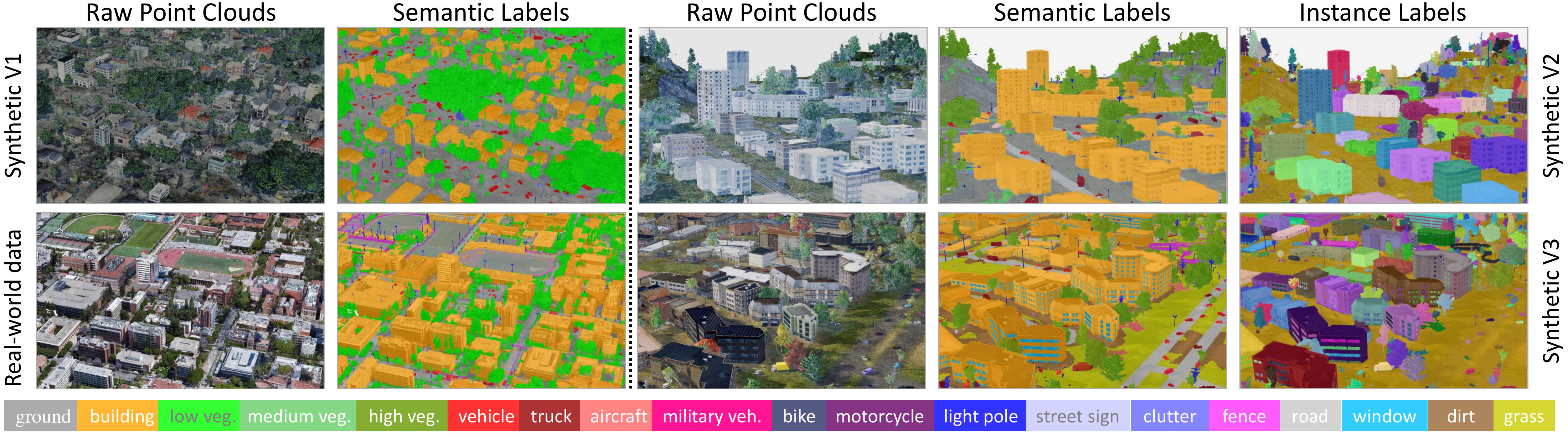}
 \vspace{-0.5cm}
	\caption{Examples of our STPLS3D dataset, including the proposed Synthetic V1, Synthetic V2, Synthetic V3, and the real-world subsets. Different semantic classes are shown in different colors, as illustrated in the color legend. Note that different instances are displayed in different random colors. Best viewed in color. }
 \vspace{-0.1cm}
	\label{fig: ExampleData}
\end{figure*}

We first conducted surveys on four real-world sites, including the University of Southern California Park Campus (USC), Wrigley Marine Science Center (WMSC) located on Catalina Island, Orange County Convention Center (OCCC), and a residential area (RA). The aerial images were collected using a crosshatch-type flight pattern with predefined overlaps ranging from 75\%$\sim$85\% and flight altitudes ranging from 25m$\sim$70m. The 3D data were reconstructed using a standard photogrammetric process and manually annotated with one of the six semantic class labels. Following that, we used our designed synthetic data generation pipeline with the same UAV flight pattern to generate an extra 62 synthetic point clouds in a wide variety of synthetic environments. In particular, three versions of the synthetic datasets were generated with different focuses. Examples of different versions of the synthetic data and real data are shown in Figure \ref{fig: ExampleData}. Please refer to Appendix \ref{sec:Data generation details} and \ref{sec:Definition of Semantic Categories} for detailed discussions of our released datasets and the available semantic labels.

\subsection{Comparison}
\label{sec:Comparison}
We provide an empirical comparison of the cost to collect the 3D data in real-world and virtual environments. Specifically, the real-world data (1.27 $km^2$) was collected with \textbf{over four months of team efforts} for data collection (including getting flight permits, planning and repeatedly executing the data collection process), processing, and annotation. By contrast, the synthetic data ($>$16 $km^2$) was generated by \textbf{a single person using one desktop PC within a month} (with an Intel Core™ i9-10900X CPU and an NVIDIA RTX 3090 with 24G memory). In particular, the time cost for synthetic data generation is not constrained by available workforce talent and can be parallel accelerated with additional computing resources.

\section{Experiments}
\label{sec:Experiments}

\subsection{Evaluation of 3D Semantic segmentation}
\label{Evaluation of 3D Semantic segmentation}

We selected five representative approaches, including PointTransformer \cite{zhao2021point}, RandLA-Net \cite{57}, SCF-Net \cite{58}, MinkowskiNet \cite{84}, and KPConv \cite{46}, as the baselines to build a semantic segmentation benchmark in our \nickname{}. Specifically, we used the original architectures of these approaches and only adapted the data-related hyperparameters to our dataset (see Appendix \ref{Data-Related Hyperparameters of Benchmark Methods}). The mean Intersection-over-Union (mIoU) and Overall Accuracy (oAcc) are used as the evaluation metrics. Note that the semantic categories of the synthetic datasets are inconsistent with the real-world dataset (18 vs. 6); see Appendix \ref{Class mapping between synthetic and real data} for details of the class mapping.

Three groups of experiments were conducted to investigate whether and how synthetic data impact the semantic segmentation performance of real-world data. Note that all three groups of experiments are tested on the test set of the real-world dataset (\ieours WMSC split), but trained with different settings: 1) Train in the real-world training set. 2) Train in synthetic datasets (V1-V3) only. 3) Train in both real and synthetic datasets.

\begin{table*}[!t]
\centering
\caption{Quantitative evaluation of the baselines on the \textit{WMSC} dataset.}
\label{tab:WMSC}
\resizebox{1.0\textwidth}{!}{%
\begin{tabular}{crcccccccc}
\Xhline{2.0\arrayrulewidth}
\multirow{2}{*}{Training sets} & \multirow{2}{*}{Methods} & \multirow{2}{*}{\textbf{mIoU (\%)}} & \multirow{2}{*}{oAcc (\%)} & \multicolumn{6}{c}{Per Class IoU (\%)}  \\ \cline{5-10} 
 & &&& Ground & Building & Tree & Car& Light pole & Fence  \\ \Xhline{2.0\arrayrulewidth}
 
\multirow{3}{*}{Real subsets} 
 & PointTransformer \cite{zhao2021point}& 36.27 & 54.31 & 39.95 & 20.88 & 62.57 & 36.13 & 49.32 & 8.76 \\
 & RandLA-Net \cite{57}& 42.33 & 60.19 & 46.13 & 24.23 & \textbf{72.46} & 53.37 & 44.82 & 12.95  \\
 & SCF-Net \cite{58}& 45.93 & \textbf{75.75} & \textbf{68.77} & \textbf{37.27} & 65.49 & 51.50 & 31.22 & \textbf{21.34} \\
   & MinkowskiNet \cite{84}& \textbf{46.52} & 70.44 & 64.22 & 29.95 & 61.33 & 45.96 & \textbf{65.25} & 12.43  \\
 & KPConv \cite{46}& 45.22 & 70.67 & 60.87 & 32.13 & 69.05 & \textbf{53.80} & 52.08 & 3.40   \\ 
 \hdashline
 
\multirow{3}{*}{Synthetic subsets} 
 & PointTransformer \cite{zhao2021point}& 45.73 & 86.76 & 84.12 & \textbf{73.37} & 60.60 & 16.96 & 27.23 & 12.10  \\
 & RandLA-Net \cite{57}& 45.03 & 81.30 & 76.78 & 57.74 & 56.08 & 28.44 & 40.36 & 10.78 \\
 & SCF-Net \cite{58}& 47.82 & 82.69 & 77.51 & 68.68 & 56.81 & 29.87 & \textbf{42.53} & 11.52  \\
   & MinkowskiNet \cite{84}& \textbf{50.78} & 87.64 & 85.23 & 72.66 & \textbf{64.80} & \textbf{31.31} & 36.85 & \textbf{13.83} \\
 & KPConv \cite{46}& 49.16 & \textbf{88.08} & \textbf{85.50} & 70.65 & 63.84 & 28.75 & 32.97 & 13.22  \\ 
 \hdashline
 
\multirow{3}{*}{Real+Synthetic}
 & PointTransformer \cite{zhao2021point}& 47.64 & 84.37 & 80.19 & 76.35 & 57.13 & 36.35 & 23.72 & 12.10 \\
 & RandLA-Net \cite{57}& 50.53 & 86.25 & 82.90 & 66.59 & 63.77 & 33.91 & 41.84 & 14.19   \\
 & SCF-Net \cite{58}& 50.65 & 83.32 & 77.80 & 58.98 & 64.86 & \textbf{46.37} & 40.50 & 15.41 \\
   & MinkowskiNet \cite{84}& 51.35 & 84.90 & 80.86 & 74.03 & 59.21 & 31.72 & \textbf{45.51} & \textbf{16.79}   \\
 & KPConv \cite{46}& \textbf{53.73} & \textbf{89.87} & \textbf{87.40} & \textbf{78.51} & \textbf{66.18} & 39.63 & 41.30 & 9.34   \\ 
 \Xhline{2.0\arrayrulewidth}
\end{tabular}%
}
\end{table*}

The quantitative performance of baselines is reported in Table \ref{tab:WMSC}. It can be seen that: 1) MinkowskiNet achieves the best overall performance with a mIoU score of 46.52\% when only trained on the real-world dataset. 2) All baselines achieved better mIoU when trained on the synthetic dataset compared with training on the real-world dataset, despite the fact that there is inevitably a domain gap that exists. This is likely because the synthetic dataset is much larger than real data in spatial scale and contains more variations of terrain shapes and building styles. 3) All baselines achieved the best mIoU when trained on real and synthetic datasets. In particular, KPConv achieved an improvement of nearly 8\% in mIoU score by training on the synthetic + real-world data. These results clearly validate that the synthetic datasets could have a positive impact on the performance of real-world 3D understanding. On the other hand, we also noticed that the addition of synthetic subsets into training sets leads to significant performance improvement for categories such as \textit{ground} and \textit{building} but with limited improvement or even worse results for small objects. This is likely due to the domain discrepancy between the real data and synthetic data. In particular, two issues need to be further addressed in future work: 1) Although we randomly assigned various materials to different objects, limited geometrical variations of 3D game objects were adopted when creating the synthetic subsets. 2) There is a lack of enforcing comprehensive contextual relationships between specific objects. For instance, cars placed off the road have random orientations in synthetic datasets, but vehicles in a parking lot are usually heading in the same direction in rows in real-world environments.

\begin{table*}[!b]
\centering
\caption{Quantitative generalization performance of baselines on the FDc dataset.}
\label{tab:FDc}
\resizebox{1.0\textwidth}{!}{%
\begin{tabular}{crcccccccc}
\Xhline{2.0\arrayrulewidth}
\multirow{2}{*}{Training sets} & \multirow{2}{*}{Methods} & \multicolumn{1}{c}{\multirow{2}{*}{\textbf{mIoU (\%)}}} & \multicolumn{1}{c}{\multirow{2}{*}{oAcc (\%)}} & \multicolumn{6}{c}{Per Class IoU (\%)}   \\ \cline{5-10} 
 & & \multicolumn{1}{c}{} & \multicolumn{1}{c}{} & \multicolumn{1}{c}{Ground} & \multicolumn{1}{c}{Building} & \multicolumn{1}{c}{Tree} & \multicolumn{1}{c}{Car} & \multicolumn{1}{c}{Light pole} & \multicolumn{1}{c}{Fence} \\ \Xhline{2.0\arrayrulewidth}

\multirow{3}{*}{Real subsets}

& PointTransformer \cite{zhao2021point} & 49.40 & 85.85 & 85.23 & 47.77 & \textbf{76.72} & 39.51& 28.61 & 18.56   \\
& RandLA-Net \cite{57} & 51.84 & 84.79 & 88.14& 46.88 & 61.40 & 48.72 & 46.04& 19.83   \\
& SCF-Net \cite{58} & 53.79 & 86.66 & \textbf{89.19} & 53.12 & 65.28 & \textbf{48.91}& 46.59 & 19.63   \\
& MinkowskiNet \cite{84} & 52.85 & 83.28 & 82.76& 40.30 & 71.68 & 47.00 & 49.33 & 26.04   \\
& KPConv \cite{46} & \textbf{57.80} & \textbf{87.20} & 86.69& \textbf{63.41} & 66.32 & 46.36 & \textbf{56.08} & \textbf{27.95}  \\ 
 
 \hdashline
\multirow{3}{*}{Synthetic subsets} 
& PointTransformer \cite{zhao2021point} & 58.65 & 92.01 & 90.42& \textbf{74.54} & 85.18 & 31.76& 42.36& 27.67   \\
& RandLA-Net \cite{57} & 59.38 & 91.33 & 90.15& 69.20 & 82.21 & 50.13 & 40.36& 24.20   \\
& SCF-Net \cite{58} & 58.82 & 90.49 & 89.53& 62.39 & 81.55 & \textbf{52.99}& 44.10& 22.36   \\
& MinkowskiNet \cite{84} & 56.17 & 90.55 & 90.74& 66.11 & 78.63 & 36.86 & 36.41& \textbf{28.26}   \\
& KPConv \cite{46} & \textbf{61.92} & \textbf{92.35} & \textbf{91.41} & 68.31 & \textbf{86.00} & 48.97 & \textbf{51.99} & 24.82  \\ 
 
 \hdashline

\multirow{3}{*}{Real+Synthetic}
& PointTransformer \cite{zhao2021point} & 62.14 & 91.96 & 89.74& \textbf{74.79} & 84.73 & 45.10& 46.75& \textbf{31.72}   \\
& RandLA-Net \cite{57} & 61.38 & 92.31 & 91.25& 68.71 & 84.35 & 55.04 & 43.30& 23.83   \\
& SCF-Net \cite{58} & 61.89 & 92.10 & 90.99& 68.69 & 84.99 & \textbf{55.58}& 45.36& 25.71   \\
& MinkowskiNet \cite{84} & 62.59 & \textbf{93.16} & 91.66& 74.70 & \textbf{87.97} & 48.80 & 43.95& 28.49   \\
& KPConv \cite{46} & \textbf{65.01} & 93.03 & \textbf{91.86} & 71.44 & 87.12 & 54.77 & \textbf{55.39} & 29.48  

 \\ \Xhline{2.0\arrayrulewidth}
\end{tabular}%
}
\end{table*}

\smallskip\noindent\textbf{Cross Datasets Generalization.} We further verified the generalization ability of the trained model on the photogrammetry Fort Drum cantonment (FDc) dataset (\ieours dataset \#7 in \cite{62}). Note that the main differences between the FDc and our STPLS3D real data are that the cold weather tree dominates the vegetation types, the aerial images were collected with smaller overlaps (50\% to 60\%), resulting in lower quality 3D data, and FDc contains various vehicle types including military vehicles that do not exist in the STPLS3D real data. Please refer to the Appendix \ref{Visualization of the FDc dataset} for data visualization.

As shown in the quantitative results reported in Table \ref{tab:FDc}, we can see that 1) KPConv consistently achieved the best generalization performance on the FDc dataset, regardless of the variation of training sets. 2) Similarly, all baselines achieved better generalization performance when trained on the synthetic dataset and achieved the best performance when trained on real and synthetic datasets. In particular, the generalization performance (\ieours mIoU score) of PointTransformer achieved an improvement of nearly 13\% when augmented with synthetic datasets during training. This clearly shows that the proposed synthetic dataset is helpful for improving the generalization capacity of the trained deep learning model.

\subsection{Evaluation of 3D Instance Segmentation}
\label{sec:Evaluation_of_3D_Instance_Segmentation}

\qy{For instance segmentation, we selected two representative voxel-based approaches, including PointGroup \cite{PointGroup} and HAIS \cite{HAIS}, as the baselines to build an instance segmentation benchmark in our \nickname{}. Considering the large spatial size of our dataset, we first tuned the data-related hyperparameters (\ieours voxel size and cluster radius) to adapt to our dataset and then utilized the weighted loss to mitigate the class imbalance issue. We followed the common practice of using the mAP, mAP50, and mAP25 as the main evaluation metrics. The SyntheticV3 dataset was selected again for evaluation, and the quantitative results achieved by different baselines are shown in Table \ref{tab:instance_segmentation}.}

\begin{table*}[b]
\setlength{\tabcolsep}{3pt}
\centering
\caption{Quantitative evaluation of two instance segmentation baselines on the synthetic v3 subset.}
\label{tab:instance_segmentation}
\resizebox{1.0\textwidth}{!}{%
\begin{tabular}{r|cc|cccccccccccccc}
\hline
 & \multicolumn{1}{l}{Metric} & \rotatebox{90}{mean (\%)}  & \rotatebox{90}{Build.} & \rotatebox{90}{LowVeg.} & \rotatebox{90}{MediumVeg.} & \rotatebox{90}{HighVeg.} & \rotatebox{90}{Vehicle} & \rotatebox{90}{Truck} & \rotatebox{90}{Aircraft} & \rotatebox{90}{MilitaryVeh.} & \rotatebox{90}{Bike} & \rotatebox{90}{Motorcycle} & \rotatebox{90}{LightPole} & \rotatebox{90}{StreetSign} & \rotatebox{90}{Clutter} & \rotatebox{90}{Fence} \\ \hline
\multirow{3}{*}{HAIS\cite{HAIS}} & AP & \textbf{35.1} & 66.8 & 20.9 & 17.6 & 23.2 & 75.7 & 51.9 & 42.6 & 31.1 & 7.4 & 50.8 & 47.0 & 8.3 & 22.6 & 25.7 \\
 & AP50 & \textbf{46.7} & 73.9 & 35.7 & 25.0 & 29.2 & 86.9 & 61.3 & 65.2 & 39.2 & 17.0 & 69.0 & 62.9 & 13.7 & 27.9 & 46.5 \\
 & AP25 & \textbf{52.8} & 75.9 & 46.8 & 31.9 & 32.1 & 89.0 & 66.0 & 72.0 & 44.5 & 22.1 & 75.4 & 68.1 & 15.0 & 31.7 & 68.4 \\ \hdashline
\multirow{3}{*}{PointGroup\cite{PointGroup}} & AP & 23.3 & 60.0 & 11.6 & 10.7 & 19.2 & 58.7 & 39.8 & 27.6 & 21.2 & 2.2 & 12.0 & 23.7 & 8.1 & 13.9 & 18.1 \\
 & AP50 & 38.5 & 70.4 & 28.3 & 19.0 & 25.4 & 83.9 & 57.9 & 47.9 & 35.3 & 7.9 & 44.0 & 46.8 & 14.7 & 19.6 & 38.4 \\
 & AP25 & 48.6 & 73.7 & 43.8 & 23.7 & 29.5 & 87.9 & 61.4 & 59.8 & 42.3 & 19.4 & 68.1 & 66.8 & 16.6 & 22.6 & 64.9 \\ \hline
\end{tabular}%
}
\end{table*}

It can be seen that HAIS outperformed PointGroup and achieved the best mAP, mAP50, and mAP25 with 40.4, 51.9, and 57.3, respectively. We also noticed that the performance of both baselines on our dataset is still far inferior to that of existing indoor datasets (\ieours Scannet \cite{9} and S3DIS \cite{8}). We attribute this performance gap to the natural differences between the large-scale outdoor and the indoor scenes, where the size of the objects in the outdoor environments are dramatically different (\ieours buildings vs. bikes) compared to the indoor scenes. In addition, the limitation of the aerial photogrammetry technique may pose extra challenges to 3D instance segmentation, where objects that are physically close to each other may not have a clear boundary in terms of geometry and texture (\ieours 3D reconstruction of forests may result in solid blobs). With the identified challenges posed by large-scale outdoor scenes, we hope our \nickname{} will pave the way for future works on designing and developing more general and effective instance segmentation techniques that can also achieve satisfactory performance on outdoor scenes.

\section{Discussions and Limitations}
 
To facilitate the research in the community, we will release not only all of the 3D point clouds but also all byproducts and relevant data, including 2D source images, annotation masks, intrinsic and extrinsic camera parameters, depth maps, and meshes. Thus \nickname{} also holds great potential for supporting other computer vision-related tasks beyond 3D semantic and instance segmentation. Tasks such as neural rendering for large outdoor scenes \cite{CityNeRF,NeRF_in_the_Wild}, style transfer for both 2D and 3D aerial data \cite{MUNIT,UNIT,DRIT++,CUT,GCGAN,CycleGAN}, 3D scene reconstruction, and object detection can all be supported.

\smallskip\noindent\textbf{Limitations.} The proposed \nickname{} has been demonstrated to have good data quality and functions; it also has limitations. First, the generated synthetic 3D environments do not have sufficient high-level contextual priors between objects, such as generating realistic site plans for houses or placing vehicles, bikes, and motorcycles in parking lots, \textit{etc.} Second, there is a visible domain gap between the synthetic and real-world data since the 2D appearance of the rendered images does not have the same style as the real-world images. We leave these domain adaptation issues for future exploration.

\section{Conclusion}
 
In this paper, we present \nickname{}, a large-scale aerial photogrammetry dataset with real and synthetic 3D point clouds. In particular, a fully automatic synthetic data generation pipeline is introduced to produce high-quality, richly-annotated 3D synthetic point clouds. Extensive experiments demonstrated the quality and functions of the generated synthetic datasets. Additionally, we also show that incorporating the synthetic data into the training set could be a good way of data augmentation, and the learning capacity and generalization ability of existing deep neural models could be further strengthened. Overall, synthetic data is easy to acquire and free of annotation, and potentially helpful for avoiding overfitting and generalized representation learning. We believe this is a promising research avenue for future research and hope our \nickname{} will inspire more research works on other tasks such as domain adaptation and pretraining.
\\
\\
\\
\noindent\textbf{Acknowledgements} The authors would like to thank the two primary sponsors of this research: US Army Simulation and Training Technology Center (STTC), and the Office of Naval Research (ONR). They would also like to acknowledge the assistance provided by Army Futures Command (AFC) and Synthetic Training Environment (STE). This work is supported by University Affiliated Research Center (UARC) award W911NF-14-D-0005. Statements and opinions expressed and content included do not necessarily reflect the position or the policy of the Government, and no official endorsement should be inferred.

\clearpage

\bibliography{egbib}

\begin{thebibliography}{86}
\providecommand{\natexlab}[1]{#1}
\providecommand{\url}[1]{\texttt{#1}}
\expandafter\ifx\csname urlstyle\endcsname\relax
  \providecommand{\doi}[1]{doi: #1}\else
  \providecommand{\doi}{doi: \begingroup \urlstyle{rm}\Url}\fi

\bibitem[Armeni et~al.(2017)Armeni, Sax, Zamir, and Savarese]{8}
Iro Armeni, Sasha Sax, Amir~R Zamir, and Silvio Savarese.
\newblock Joint 2d-3d-semantic data for indoor scene understanding.
\newblock In \emph{CVPR}, 2017.

\bibitem[ASPRS(2013)]{61}
ASPRS.
\newblock Las specification 1.4-r14, 2013.

\bibitem[Behley et~al.(2019)Behley, Garbade, Milioto, Quenzel, Behnke,
  Stachniss, and Gall]{15}
Jens Behley, Martin Garbade, Andres Milioto, Jan Quenzel, Sven Behnke, Cyrill
  Stachniss, and Jurgen Gall.
\newblock Semantickitti: A dataset for semantic scene understanding of lidar
  sequences.
\newblock In \emph{ICCV}, 2019.

\bibitem[Benz et~al.(2021)Benz, Taraben, Debus, Habte, Oppermann, Hallermann,
  Voelker, Rodehorst, and Morgenthal]{34}
Alexander Benz, Jakob Taraben, Paul Debus, Bedilu Habte, Luise Oppermann,
  Norman Hallermann, Conrad Voelker, Volker Rodehorst, and Guido Morgenthal.
\newblock Framework for a uas-based assessment of energy performance of
  buildings.
\newblock \emph{Energy and Buildings}, 2021.

\bibitem[Caesar et~al.(2020)Caesar, Bankiti, Lang, Vora, Liong, Xu, Krishnan,
  Pan, Baldan, and Beijbom]{81}
Holger Caesar, Varun Bankiti, Alex~H Lang, Sourabh Vora, Venice~Erin Liong,
  Qiang Xu, Anush Krishnan, Yu~Pan, Giancarlo Baldan, and Oscar Beijbom.
\newblock nuscenes: A multimodal dataset for autonomous driving.
\newblock In \emph{CVPR}, 2020.

\bibitem[Can et~al.(2021)Can, Mantegazza, Abbate, Chappuis, and Giusti]{25}
G{\"u}lcan Can, Dario Mantegazza, Gabriele Abbate, S{\'e}bastien Chappuis, and
  Alessandro Giusti.
\newblock Semantic segmentation on swiss3dcities: A benchmark study on aerial
  photogrammetric 3d pointcloud dataset.
\newblock \emph{Pattern Recognition Letters}, 2021.

\bibitem[Chang et~al.(2017)Chang, Dai, Funkhouser, Halber, Niessner, Savva,
  Song, Zeng, and Zhang]{7}
Angel Chang, Angela Dai, Thomas Funkhouser, Maciej Halber, Matthias Niessner,
  Manolis Savva, Shuran Song, Andy Zeng, and Yinda Zhang.
\newblock {Matterport3D}: Learning from {RGB-D} data in indoor environments.
\newblock In \emph{3DV}, 2017.

\bibitem[Chang et~al.(2015)Chang, Funkhouser, Guibas, Hanrahan, Huang, Li,
  Savarese, Savva, Song, Su, Xiao, Yi, and Yu]{4}
Angel~X. Chang, Thomas Funkhouser, Leonidas Guibas, Pat Hanrahan, Qixing Huang,
  Zimo Li, Silvio Savarese, Manolis Savva, Shuran Song, Hao Su, Jianxiong Xiao,
  Li~Yi, and Fisher Yu.
\newblock {ShapeNet: An Information-Rich 3D Model Repository}.
\newblock Technical Report arXiv:1512.03012 [cs.GR], 2015.

\bibitem[Chang et~al.(2019)Chang, Lambert, Sangkloy, Singh, Bak, Hartnett,
  Wang, Carr, Lucey, Ramanan, et~al.]{79}
Ming-Fang Chang, John Lambert, Patsorn Sangkloy, Jagjeet Singh, Slawomir Bak,
  Andrew Hartnett, De~Wang, Peter Carr, Simon Lucey, Deva Ramanan, et~al.
\newblock Argoverse: 3d tracking and forecasting with rich maps.
\newblock In \emph{CVPR}, 2019.

\bibitem[Chen et~al.(2020{\natexlab{a}})Chen, Feng, McAlinden, and
  Soibelman]{chen2020photogrammetric}
Meida Chen, Andrew Feng, Ryan McAlinden, and Lucio Soibelman.
\newblock Photogrammetric point cloud segmentation and object information
  extraction for creating virtual environments and simulations.
\newblock \emph{Journal of Management in Engineering}, 36\penalty0
  (2):\penalty0 04019046, 2020{\natexlab{a}}.

\bibitem[Chen et~al.(2020{\natexlab{b}})Chen, Feng, McCullough, Prasad,
  McAlinden, and Soibelman]{62}
Meida Chen, Andrew Feng, Kyle McCullough, Pratusha~Bhuvana Prasad, Ryan
  McAlinden, and Lucio Soibelman.
\newblock 3d photogrammetry point cloud segmentation using a model ensembling
  framework.
\newblock \emph{ASCE JCCE}, 2020{\natexlab{b}}.

\bibitem[Chen et~al.(2020{\natexlab{c}})Chen, Feng, McCullough, Prasad,
  McAlinden, and Soibelman]{chen2020generating}
Meida Chen, Andrew Feng, Kyle McCullough, Pratusha~Bhuvana Prasad, Ryan
  McAlinden, and Lucio Soibelman.
\newblock Generating synthetic photogrammetric data for training deep learning
  based 3d point cloud segmentation models.
\newblock In \emph{I/ITSEC}, 2020{\natexlab{c}}.

\bibitem[Chen et~al.(2020{\natexlab{d}})Chen, Feng, McCullough, Prasad,
  McAlinden, and Soibelman]{chen2020semantic}
Meida Chen, Andrew Feng, Kyle McCullough, Pratusha~Bhuvana Prasad, Ryan
  McAlinden, and Lucio Soibelman.
\newblock Semantic segmentation and data fusion of microsoft bing 3d cities and
  small uav-based photogrammetric data.
\newblock In \emph{I/ITSEC}, 2020{\natexlab{d}}.

\bibitem[Chen et~al.(2020{\natexlab{e}})Chen, Feng, McCullough, Prasad,
  McAlinden, Soibelman, and Enloe]{chen2020fully}
Meida Chen, Andrew Feng, Kyle McCullough, Pratusha~Bhuvana Prasad, Ryan
  McAlinden, Lucio Soibelman, and Mike Enloe.
\newblock Fully automated photogrammetric data segmentation and object
  information extraction approach for creating simulation terrain.
\newblock In \emph{I/ITSEC}, 2020{\natexlab{e}}.

\bibitem[Chen et~al.(2021{\natexlab{a}})Chen, Feng, Hou, McCullough, Prasad,
  and Soibelman]{56}
Meida Chen, Andrew Feng, Yu~Hou, Kyle McCullough, Pratusha~Bhuvana Prasad, and
  Lucio Soibelman.
\newblock Ground material classification and for uav-based photogrammetric 3d
  data a 2d-3d hybrid approach.
\newblock In \emph{I/ITSEC}, 2021{\natexlab{a}}.

\bibitem[Chen et~al.(2021{\natexlab{b}})Chen, Fang, Zhang, Liu, and Wang]{HAIS}
Shaoyu Chen, Jiemin Fang, Qian Zhang, Wenyu Liu, and Xinggang Wang.
\newblock Hierarchical aggregation for 3d instance segmentation.
\newblock In \emph{ICCV}, pages 15467--15476, 2021{\natexlab{b}}.

\bibitem[Choy et~al.(2019)Choy, Gwak, and Savarese]{84}
Christopher Choy, JunYoung Gwak, and Silvio Savarese.
\newblock 4d spatio-temporal convnets: Minkowski convolutional neural networks.
\newblock In \emph{CVPR}, 2019.

\bibitem[Dai et~al.(2017)Dai, Chang, Savva, Halber, Funkhouser, and
  Nie{\ss}ner]{9}
Angela Dai, Angel~X Chang, Manolis Savva, Maciej Halber, Thomas Funkhouser, and
  Matthias Nie{\ss}ner.
\newblock Scannet: Richly-annotated 3d reconstructions of indoor scenes.
\newblock In \emph{CVPR}, 2017.

\bibitem[De~Deuge et~al.(2013)De~Deuge, Quadros, Hung, and Douillard]{10}
Mark De~Deuge, Alastair Quadros, Calvin Hung, and Bertrand Douillard.
\newblock Unsupervised feature learning for classification of outdoor 3d scans.
\newblock In \emph{ACRA}, 2013.

\bibitem[Dosovitskiy et~al.(2017)Dosovitskiy, Ros, Codevilla, Lopez, and
  Koltun]{70}
Alexey Dosovitskiy, German Ros, Felipe Codevilla, Antonio Lopez, and Vladlen
  Koltun.
\newblock {CARLA}: {An} open urban driving simulator.
\newblock In \emph{CoRL}, 2017.

\bibitem[Erikson(2003)]{erikson2003segmentation}
Mats Erikson.
\newblock Segmentation of individual tree crowns in colour aerial photographs
  using region growing supported by fuzzy rules.
\newblock \emph{Canadian Journal of Forest Research}, 2003.

\bibitem[Fan et~al.(2021)Fan, Dong, Zhu, Lv, Ye, and Wang]{58}
Siqi Fan, Qiulei Dong, Fenghua Zhu, Yisheng Lv, Peijun Ye, and Fei-Yue Wang.
\newblock Scf-net: Learning spatial contextual features for large-scale point
  cloud segmentation.
\newblock In \emph{CVPR}, 2021.

\bibitem[Fu et~al.(2019)Fu, Gong, Wang, Batmanghelich, Zhang, and Tao]{GCGAN}
Huan Fu, Mingming Gong, Chaohui Wang, Kayhan Batmanghelich, Kun Zhang, and
  Dacheng Tao.
\newblock Geometry-consistent generative adversarial networks for one-sided
  unsupervised domain mapping.
\newblock In \emph{CVPR}, pages 2427--2436, 2019.

\bibitem[Gaidon et~al.(2016)Gaidon, Wang, Cabon, and Vig]{63}
Adrien Gaidon, Qiao Wang, Yohann Cabon, and Eleonora Vig.
\newblock Virtual worlds as proxy for multi-object tracking analysis.
\newblock In \emph{CVPR}, 2016.

\bibitem[Gao et~al.(2021)Gao, Pan, Li, Geng, and Zhao]{2}
Biao Gao, Yancheng Pan, Chengkun Li, Sibo Geng, and Huijing Zhao.
\newblock Are we hungry for 3d lidar data for semantic segmentation? a survey
  of datasets and methods.
\newblock \emph{IEEE Transactions on ITS}, 2021.

\bibitem[Gesch et~al.(2009)Gesch, Evans, Mauck, Hutchinson, Carswell~Jr,
  et~al.]{68}
Dean Gesch, Gayla Evans, James Mauck, John Hutchinson, William~J Carswell~Jr,
  et~al.
\newblock The national map—elevation.
\newblock \emph{US geological survey fact sheet}, 2009.

\bibitem[Geyer et~al.(2020)Geyer, Kassahun, Mahmudi, Ricou, Durgesh, Chung,
  Hauswald, Pham, M{\"u}hlegg, Dorn, et~al.]{22}
Jakob Geyer, Yohannes Kassahun, Mentar Mahmudi, Xavier Ricou, Rupesh Durgesh,
  Andrew~S Chung, Lorenz Hauswald, Viet~Hoang Pham, Maximilian M{\"u}hlegg,
  Sebastian Dorn, et~al.
\newblock A2d2: Audi autonomous driving dataset.
\newblock \emph{arXiv preprint arXiv:2004.06320}, 2020.

\bibitem[Griffiths and Boehm(2019)]{26}
David Griffiths and Jan Boehm.
\newblock Synthcity: A large scale synthetic point cloud.
\newblock \emph{arXiv preprint arXiv:1907.04758}, 2019.

\bibitem[Guo et~al.(2020)Guo, Wang, Hu, Liu, Liu, and Bennamoun]{guo2020deep}
Yulan Guo, Hanyun Wang, Qingyong Hu, Hao Liu, Li~Liu, and Mohammed Bennamoun.
\newblock Deep learning for 3d point clouds: A survey.
\newblock \emph{IEEE transactions on pattern analysis and machine
  intelligence}, 43\penalty0 (12):\penalty0 4338--4364, 2020.

\bibitem[Hackel et~al.(2017)Hackel, Savinov, Ladicky, Wegner, Schindler, and
  Pollefeys]{3}
Timo Hackel, Nikolay Savinov, Lubor Ladicky, Jan~D. Wegner, Konrad Schindler,
  and Marc Pollefeys.
\newblock {SEMANTIC3D.NET: A new large-scale point cloud classification
  benchmark}.
\newblock In \emph{ISPRS Journal of Photogrammetry and Remote Sensing}, 2017.

\bibitem[Haklay and Weber(2008)]{69}
Mordechai Haklay and Patrick Weber.
\newblock Openstreetmap: User-generated street maps.
\newblock \emph{IEEE Pervasive computing}, 2008.

\bibitem[Ham et~al.(2016)Ham, Han, Lin, and Golparvar-Fard]{31}
Youngjib Ham, Kevin~K Han, Jacob~J Lin, and Mani Golparvar-Fard.
\newblock Visual monitoring of civil infrastructure systems via camera-equipped
  unmanned aerial vehicles (uavs): a review of related works.
\newblock \emph{Visualization in Engineering}, 2016.

\bibitem[Hou et~al.(2021{\natexlab{a}})Hou, Chen, Volk, and Soibelman]{32}
Yu~Hou, Meida Chen, Rebekka Volk, and Lucio Soibelman.
\newblock Investigation on performance of rgb point cloud and thermal
  information data fusion for building thermal map modeling using aerial images
  under different experimental conditions.
\newblock \emph{JOBE}, 2021{\natexlab{a}}.

\bibitem[Hou et~al.(2021{\natexlab{b}})Hou, Chen, Volk, and
  Soibelman]{hou2021approach}
Yu~Hou, Meida Chen, Rebekka Volk, and Lucio Soibelman.
\newblock An approach to semantically segmenting building components and
  outdoor scenes based on multichannel aerial imagery datasets.
\newblock \emph{Remote Sensing}, 13\penalty0 (21):\penalty0 4357,
  2021{\natexlab{b}}.

\bibitem[Hu et~al.(2020)Hu, Yang, Xie, Rosa, Guo, Wang, Trigoni, and
  Markham]{57}
Qingyong Hu, Bo~Yang, Linhai Xie, Stefano Rosa, Yulan Guo, Zhihua Wang, Niki
  Trigoni, and Andrew Markham.
\newblock Randla-net: Efficient semantic segmentation of large-scale point
  clouds.
\newblock In \emph{CVPR}, 2020.

\bibitem[Hu et~al.(2021{\natexlab{a}})Hu, Yang, Khalid, Xiao, Trigoni, and
  Markham]{23}
Qingyong Hu, Bo~Yang, Sheikh Khalid, Wen Xiao, Niki Trigoni, and Andrew
  Markham.
\newblock Towards semantic segmentation of urban-scale 3d point clouds: A
  dataset, benchmarks and challenges.
\newblock In \emph{CVPR}, 2021{\natexlab{a}}.

\bibitem[Hu et~al.(2021{\natexlab{b}})Hu, Yang, Xie, Rosa, Guo, Wang, Trigoni,
  and Markham]{hu2021learning}
Qingyong Hu, Bo~Yang, Linhai Xie, Stefano Rosa, Yulan Guo, Zhihua Wang, Niki
  Trigoni, and Andrew Markham.
\newblock Learning semantic segmentation of large-scale point clouds with
  random sampling.
\newblock \emph{IEEE Transactions on Pattern Analysis and Machine
  Intelligence}, 2021{\natexlab{b}}.

\bibitem[Hu et~al.(2022{\natexlab{a}})Hu, Yang, Fang, Guo, Leonardis, Trigoni,
  and Markham]{hu2021sqn}
Qingyong Hu, Bo~Yang, Guangchi Fang, Yulan Guo, Ales Leonardis, Niki Trigoni,
  and Andrew Markham.
\newblock Sqn: Weakly-supervised semantic segmentation of large-scale 3d point
  clouds.
\newblock In \emph{ECCV}, 2022{\natexlab{a}}.

\bibitem[Hu et~al.(2022{\natexlab{b}})Hu, Yang, Khalid, Xiao, Trigoni, and
  Markham]{hu2022sensaturban}
Qingyong Hu, Bo~Yang, Sheikh Khalid, Wen Xiao, Niki Trigoni, and Andrew
  Markham.
\newblock Sensaturban: Learning semantics from urban-scale photogrammetric
  point clouds.
\newblock \emph{International Journal of Computer Vision}, pages 1--28,
  2022{\natexlab{b}}.

\bibitem[Huang et~al.(2018)Huang, Liu, Belongie, and Kautz]{MUNIT}
Xun Huang, Ming-Yu Liu, Serge Belongie, and Jan Kautz.
\newblock Multimodal unsupervised image-to-image translation.
\newblock In \emph{ECCV}, pages 172--189, 2018.

\bibitem[Jiang et~al.(2020)Jiang, Zhao, Shi, Liu, Fu, and Jia]{PointGroup}
Li~Jiang, Hengshuang Zhao, Shaoshuai Shi, Shu Liu, Chi-Wing Fu, and Jiaya Jia.
\newblock Pointgroup: Dual-set point grouping for 3d instance segmentation.
\newblock In \emph{CVPR}, pages 4867--4876, 2020.

\bibitem[Johnson-Roberson et~al.(2017)Johnson-Roberson, Barto, Mehta, Sridhar,
  Rosaen, and Vasudevan]{71}
Matthew Johnson-Roberson, Charles Barto, Rounak Mehta, Sharath~Nittur Sridhar,
  Karl Rosaen, and Ram Vasudevan.
\newblock Driving in the matrix: Can virtual worlds replace human-generated
  annotations for real world tasks?
\newblock In \emph{ICRA}, 2017.

\bibitem[K{\"o}lle et~al.(2021)K{\"o}lle, Laupheimer, Schmohl, Haala,
  Rottensteiner, Wegner, and Ledoux]{KOLLE2021100001}
Michael K{\"o}lle, Dominik Laupheimer, Stefan Schmohl, Norbert Haala, Franz
  Rottensteiner, Jan~Dirk Wegner, and Hugo Ledoux.
\newblock The hessigheim 3d (h3d) benchmark on semantic segmentation of
  high-resolution 3d point clouds and textured meshes from uav lidar and
  multi-view-stereo.
\newblock \emph{ISPRS Open Journal of Photogrammetry and Remote Sensing},
  1:\penalty0 100001, 2021.

\bibitem[Landrieu and Simonovsky(2018)]{54}
Loic Landrieu and Martin Simonovsky.
\newblock Large-scale point cloud semantic segmentation with superpoint graphs.
\newblock In \emph{CVPR}, 2018.

\bibitem[Le et~al.(2021)Le, Das, Mensink, Karaoglu, and Gevers]{le21wacv}
Hoang{-}An Le, Partha Das, Thomas Mensink, Sezer Karaoglu, and Theo Gevers.
\newblock {EDEN: Multimodal Synthetic Dataset of Enclosed garDEN Scenes}.
\newblock In \emph{WACV}, 2021.

\bibitem[Lee et~al.(2018)Lee, Tseng, Huang, Singh, and Yang]{DRIT++}
Hsin-Ying Lee, Hung-Yu Tseng, Jia-Bin Huang, Maneesh Singh, and Ming-Hsuan
  Yang.
\newblock Diverse image-to-image translation via disentangled representations.
\newblock In \emph{ECCV}, pages 35--51, 2018.

\bibitem[Li et~al.(2017)Li, Fu, Yu, and Cracknell]{li2017deep}
Weijia Li, Haohuan Fu, Le~Yu, and Arthur Cracknell.
\newblock Deep learning based oil palm tree detection and counting for
  high-resolution remote sensing images.
\newblock \emph{Remote Sensing}, 2017.

\bibitem[Li et~al.(2020)Li, Li, Tong, Lim, Yuan, Wu, Tang, and Huang]{24}
Xinke Li, Chongshou Li, Zekun Tong, Andrew Lim, Junsong Yuan, Yuwei Wu, Jing
  Tang, and Raymond Huang.
\newblock Campus3d: A photogrammetry point cloud benchmark for hierarchical
  understanding of outdoor scene.
\newblock In \emph{ACM MM}, 2020.

\bibitem[Liu et~al.(2017)Liu, Breuel, and Kautz]{UNIT}
Ming-Yu Liu, Thomas Breuel, and Jan Kautz.
\newblock Unsupervised image-to-image translation networks.
\newblock In \emph{Advances in neural information processing systems}, pages
  700--708, 2017.

\bibitem[L{\'o}pez et~al.(2021)L{\'o}pez, Jurado, Ogayar, and Feito]{33}
Alfonso L{\'o}pez, Juan~M Jurado, Carlos~J Ogayar, and Francisco~R Feito.
\newblock An optimized approach for generating dense thermal point clouds from
  uav-imagery.
\newblock \emph{ISPRS Journal of Photogrammetry and Remote Sensing}, 2021.

\bibitem[Martin-Brualla et~al.(2021)Martin-Brualla, Radwan, Sajjadi, Barron,
  Dosovitskiy, and Duckworth]{NeRF_in_the_Wild}
Ricardo Martin-Brualla, Noha Radwan, Mehdi S.~M. Sajjadi, Jonathan~T. Barron,
  Alexey Dosovitskiy, and Daniel Duckworth.
\newblock Nerf in the wild: Neural radiance fields for unconstrained photo
  collections.
\newblock In \emph{CVPR}, 2021.

\bibitem[Mo et~al.(2019)Mo, Zhu, Chang, Yi, Tripathi, Guibas, and Su]{5}
Kaichun Mo, Shilin Zhu, Angel~X Chang, Li~Yi, Subarna Tripathi, Leonidas~J
  Guibas, and Hao Su.
\newblock Partnet: A large-scale benchmark for fine-grained and hierarchical
  part-level 3d object understanding.
\newblock In \emph{CVPR}, 2019.

\bibitem[Munoz et~al.(2009{\natexlab{a}})Munoz, Bagnell, Vandapel, and
  Hebert]{11}
Daniel Munoz, J~Andrew Bagnell, Nicolas Vandapel, and Martial Hebert.
\newblock Contextual classification with functional max-margin markov networks.
\newblock In \emph{CVPR}, 2009{\natexlab{a}}.

\bibitem[Munoz et~al.(2009{\natexlab{b}})Munoz, Bagnell, Vandapel, and
  Hebert]{77}
Daniel Munoz, J~Andrew Bagnell, Nicolas Vandapel, and Martial Hebert.
\newblock Contextual classification with functional max-margin markov networks.
\newblock In \emph{CVPR}, 2009{\natexlab{b}}.

\bibitem[Pan et~al.(2020)Pan, Gao, Mei, Geng, Li, and Zhao]{80}
Yancheng Pan, Biao Gao, Jilin Mei, Sibo Geng, Chengkun Li, and Huijing Zhao.
\newblock Semanticposs: A point cloud dataset with large quantity of dynamic
  instances.
\newblock In \emph{IEEE IV}, 2020.

\bibitem[Park et~al.(2020)Park, Efros, Zhang, and Zhu]{CUT}
Taesung Park, Alexei~A Efros, Richard Zhang, and Jun-Yan Zhu.
\newblock Contrastive learning for unpaired image-to-image translation.
\newblock In \emph{ECCV}, pages 319--345, 2020.

\bibitem[Peng et~al.(2020)Peng, Usman, Saito, Kaushik, Hoffman, and Saenko]{65}
Xingchao Peng, Ben Usman, Kuniaki Saito, Neela Kaushik, Judy Hoffman, and Kate
  Saenko.
\newblock Syn2real: A new benchmark for synthetic-to-real visual domain
  adaptation.
\newblock In \emph{CVPR}, 2020.

\bibitem[Qiu et~al.(2021)Qiu, Anwar, and Barnes]{94}
Shi Qiu, Saeed Anwar, and Nick Barnes.
\newblock Semantic segmentation for real point cloud scenes via bilateral
  augmentation and adaptive fusion.
\newblock In \emph{CVPR}, 2021.

\bibitem[Richter et~al.(2016)Richter, Vineet, Roth, and Koltun]{1}
Stephan~R Richter, Vibhav Vineet, Stefan Roth, and Vladlen Koltun.
\newblock Playing for data: Ground truth from computer games.
\newblock In \emph{ECCV}, 2016.

\bibitem[Ros et~al.(2016)Ros, Sellart, Materzynska, Vazquez, and Lopez]{64}
German Ros, Laura Sellart, Joanna Materzynska, David Vazquez, and Antonio~M
  Lopez.
\newblock The synthia dataset: A large collection of synthetic images for
  semantic segmentation of urban scenes.
\newblock In \emph{CVPR}, 2016.

\bibitem[Rottensteiner et~al.(2012)Rottensteiner, Sohn, Jung, Gerke, Baillard,
  Benitez, and Breitkopf]{17}
Franz Rottensteiner, Gunho Sohn, Jaewook Jung, Markus Gerke, Caroline Baillard,
  Sebastien Benitez, and Uwe Breitkopf.
\newblock The isprs benchmark on urban object classification and 3d building
  reconstruction.
\newblock \emph{ISPRS Journal of Photogrammetry and Remote Sensing}, 2012.

\bibitem[Roynard et~al.(2018)Roynard, Deschaud, and Goulette]{14}
Xavier Roynard, Jean-Emmanuel Deschaud, and Fran{\c{c}}ois Goulette.
\newblock Paris-lille-3d: A large and high-quality ground-truth urban point
  cloud dataset for automatic segmentation and classification.
\newblock \emph{IJRR}, 2018.

\bibitem[Scharw{\"a}chter et~al.(2013)Scharw{\"a}chter, Enzweiler, Franke, and
  Roth]{21}
Timo Scharw{\"a}chter, Markus Enzweiler, Uwe Franke, and Stefan Roth.
\newblock Efficient multi-cue scene segmentation.
\newblock In \emph{GCPR}, 2013.

\bibitem[Serna et~al.(2014)Serna, Marcotegui, Goulette, and Deschaud]{12}
Andr{\'e}s Serna, Beatriz Marcotegui, Fran{\c{c}}ois Goulette, and
  Jean-Emmanuel Deschaud.
\newblock Paris-rue-madame database: a 3d mobile laser scanner dataset for
  benchmarking urban detection, segmentation and classification methods.
\newblock In \emph{ICPRAM}, 2014.

\bibitem[Shah et~al.(2018)Shah, Dey, Lovett, and Kapoor]{59}
Shital Shah, Debadeepta Dey, Chris Lovett, and Ashish Kapoor.
\newblock Airsim: High-fidelity visual and physical simulation for autonomous
  vehicles.
\newblock In \emph{Field and service robotics}, 2018.

\bibitem[Shi et~al.(2021)Shi, Kang, Xia, Tyagi, Mehta, and Du]{shi2021spatial}
Yangming Shi, John Kang, Pengxiang Xia, Oshin Tyagi, Ranjana~K Mehta, and Jing
  Du.
\newblock Spatial knowledge and firefighters’ wayfinding performance: A
  virtual reality search and rescue experiment.
\newblock \emph{Safety science}, 139:\penalty0 105231, 2021.

\bibitem[Silberman et~al.(2012)Silberman, Hoiem, Kohli, and Fergus]{6}
Nathan Silberman, Derek Hoiem, Pushmeet Kohli, and Rob Fergus.
\newblock Indoor segmentation and support inference from rgbd images.
\newblock In \emph{ECCV}, 2012.

\bibitem[Song et~al.(2015)Song, Lichtenberg, and Xiao]{76}
Shuran Song, Samuel~P. Lichtenberg, and Jianxiong Xiao.
\newblock Sun rgb-d: A rgb-d scene understanding benchmark suite.
\newblock In \emph{CVPR}, 2015.

\bibitem[Spicer et~al.(2016)Spicer, McAlinden, Conover, and Adelphi]{60}
Ryan Spicer, Ryan McAlinden, Damon Conover, and M~Adelphi.
\newblock Producing usable simulation terrain data from uas-collected imagery.
\newblock In \emph{I/ITSEC}, 2016.

\bibitem[Sun et~al.(2020)Sun, Kretzschmar, Dotiwalla, Chouard, Patnaik, Tsui,
  Guo, Zhou, Chai, Caine, Vasudevan, Han, Ngiam, Zhao, Timofeev, Ettinger,
  Krivokon, Gao, Joshi, Zhang, Shlens, Chen, and Anguelov]{83}
Pei Sun, Henrik Kretzschmar, Xerxes Dotiwalla, Aurelien Chouard, Vijaysai
  Patnaik, Paul Tsui, James Guo, Yin Zhou, Yuning Chai, Benjamin Caine, Vijay
  Vasudevan, Wei Han, Jiquan Ngiam, Hang Zhao, Aleksei Timofeev, Scott
  Ettinger, Maxim Krivokon, Amy Gao, Aditya Joshi, Yu~Zhang, Jonathon Shlens,
  Zhifeng Chen, and Dragomir Anguelov.
\newblock Scalability in perception for autonomous driving: Waymo open dataset.
\newblock In \emph{CVPR}, 2020.

\bibitem[Tan et~al.(2020)Tan, Qin, Ma, Li, Du, Cai, Yang, and Li]{16}
Weikai Tan, Nannan Qin, Lingfei Ma, Ying Li, Jing Du, Guorong Cai, Ke~Yang, and
  Jonathan Li.
\newblock Toronto-3d: A large-scale mobile lidar dataset for semantic
  segmentation of urban roadways.
\newblock In \emph{CVPR}, 2020.

\bibitem[Thomas et~al.(2019)Thomas, Qi, Deschaud, Marcotegui, Goulette, and
  Guibas]{46}
Hugues Thomas, Charles~R. Qi, Jean-Emmanuel Deschaud, Beatriz Marcotegui,
  Francois Goulette, and Leonidas~J. Guibas.
\newblock Kpconv: Flexible and deformable convolution for point clouds.
\newblock In \emph{ICCV}, 2019.

\bibitem[Tong et~al.(2020)Tong, Li, Chen, Sun, Cao, and Xiang]{82}
Guofeng Tong, Yong Li, Dong Chen, Qi~Sun, Wei Cao, and Guiqiu Xiang.
\newblock Cspc-dataset: new lidar point cloud dataset and benchmark for
  large-scale scene semantic segmentation.
\newblock \emph{IEEE Access}, 2020.

\bibitem[Uy et~al.(2019)Uy, Pham, Hua, Nguyen, and Yeung]{75}
Mikaela~Angelina Uy, Quang-Hieu Pham, Binh-Son Hua, Thanh Nguyen, and Sai-Kit
  Yeung.
\newblock Revisiting point cloud classification: A new benchmark dataset and
  classification model on real-world data.
\newblock In \emph{ICCV}, 2019.

\bibitem[Vallet et~al.(2015)Vallet, Br{\'e}dif, Serna, Marcotegui, and
  Paparoditis]{13}
Bruno Vallet, Mathieu Br{\'e}dif, Andr{\'e}s Serna, Beatriz Marcotegui, and
  Nicolas Paparoditis.
\newblock Terramobilita/iqmulus urban point cloud analysis benchmark.
\newblock \emph{Computers \& Graphics}, 2015.

\bibitem[Varney et~al.(2020)Varney, Asari, and Graehling]{19}
Nina Varney, Vijayan~K Asari, and Quinn Graehling.
\newblock Dales: a large-scale aerial lidar data set for semantic segmentation.
\newblock In \emph{CVPRW}, 2020.

\bibitem[Wu et~al.(2015)Wu, Song, Khosla, Yu, Zhang, Tang, and Xiao]{73}
Zhirong Wu, Shuran Song, Aditya Khosla, Fisher Yu, Linguang Zhang, Xiaoou Tang,
  and Jianxiong Xiao.
\newblock 3d shapenets: A deep representation for volumetric shapes.
\newblock In \emph{CVPR}, 2015.

\bibitem[Xiangli et~al.(2021)Xiangli, Xu, Pan, Zhao, Rao, Theobalt, Dai, and
  Lin]{CityNeRF}
Yuanbo Xiangli, Linning Xu, Xingang Pan, Nanxuan Zhao, Anyi Rao, Christian
  Theobalt, Bo~Dai, and Dahua Lin.
\newblock Citynerf: Building nerf at city scale.
\newblock \emph{arXiv preprint arXiv:2112.05504}, 2021.

\bibitem[Xiao et~al.(2021)Xiao, Huang, Guan, Zhan, and Lu]{SynLiDAR}
Aoran Xiao, Jiaxing Huang, Dayan Guan, Fangneng Zhan, and Shijian Lu.
\newblock Synlidar: Learning from synthetic lidar sequential point cloud for
  semantic segmentation.
\newblock \emph{arXiv preprint arXiv:2107.05399}, 2021.

\bibitem[Yang et~al.(2015)Yang, Park, Vela, and Golparvar-Fard]{30}
Jun Yang, Man-Woo Park, Patricio~A Vela, and Mani Golparvar-Fard.
\newblock Construction performance monitoring via still images, time-lapse
  photos, and video streams: Now, tomorrow, and the future.
\newblock \emph{Advanced Engineering Informatics}, 2015.

\bibitem[Ye et~al.(2020)Ye, Xu, Huang, Tong, Li, Liu, Luan, Hoegner, and
  Stilla]{20}
Zhen Ye, Yusheng Xu, Rong Huang, Xiaohua Tong, Xin Li, Xiangfeng Liu, Kuifeng
  Luan, Ludwig Hoegner, and Uwe Stilla.
\newblock Lasdu: A large-scale aerial lidar dataset for semantic labeling in
  dense urban areas.
\newblock \emph{ISPRS International Journal of Geo-Information}, 2020.

\bibitem[Yi et~al.(2016)Yi, Kim, Ceylan, Shen, Yan, Su, Lu, Huang, Sheffer, and
  Guibas]{74}
Li~Yi, Vladimir~G Kim, Duygu Ceylan, I-Chao Shen, Mengyan Yan, Hao Su, Cewu Lu,
  Qixing Huang, Alla Sheffer, and Leonidas Guibas.
\newblock A scalable active framework for region annotation in 3d shape
  collections.
\newblock \emph{ACM TOG}, 2016.

\bibitem[Yue et~al.(2018)Yue, Wu, Seshia, Keutzer, and
  Sangiovanni-Vincentelli]{72}
Xiangyu Yue, Bichen Wu, Sanjit~A Seshia, Kurt Keutzer, and Alberto~L
  Sangiovanni-Vincentelli.
\newblock A lidar point cloud generator: from a virtual world to autonomous
  driving.
\newblock In \emph{ACM ICMR}, 2018.

\bibitem[Zhao et~al.(2021)Zhao, Jiang, Jia, Torr, and Koltun]{zhao2021point}
Hengshuang Zhao, Li~Jiang, Jiaya Jia, Philip~HS Torr, and Vladlen Koltun.
\newblock Point transformer.
\newblock In \emph{ICCV}, pages 16259--16268, 2021.

\bibitem[Zhu et~al.(2017)Zhu, Park, Isola, and Efros]{CycleGAN}
Jun-Yan Zhu, Taesung Park, Phillip Isola, and Alexei~A Efros.
\newblock Unpaired image-to-image translation using cycle-consistent
  adversarial networks.
\newblock In \emph{ICCV}, pages 2223--2232, 2017.

\bibitem[Zolanvari et~al.(2019)Zolanvari, Ruano, Rana, Cummins, da~Silva,
  Rahbar, and Smolic]{18}
SM~Zolanvari, Susana Ruano, Aakanksha Rana, Alan Cummins, Rogerio~Eduardo
  da~Silva, Morteza Rahbar, and Aljosa Smolic.
\newblock Dublincity: Annotated lidar point cloud and its applications.
\newblock In \emph{BMVC}, 2019.

\end{thebibliography}

\clearpage

\appendix
\section*{Appendices}
\addcontentsline{toc}{section}{Appendices}
\renewcommand{\thesubsection}{\Alph{subsection}}





\section{Data availability}

Our STPLS3D datasets can be downloaded at: \dataDownload

\section{Data collection \& generation details}
\label{sec:Data generation details}

\textbf{Video Illustrations.} \qy{We provide an video demo illustrating our synthetic data generation pipeline discussed in Section \ref{section:3}. The video can be viewed at \noindent\workflowVideo.} 

\begin{figure}[b]
	\begin{center}
	\includegraphics[width=0.6\linewidth]{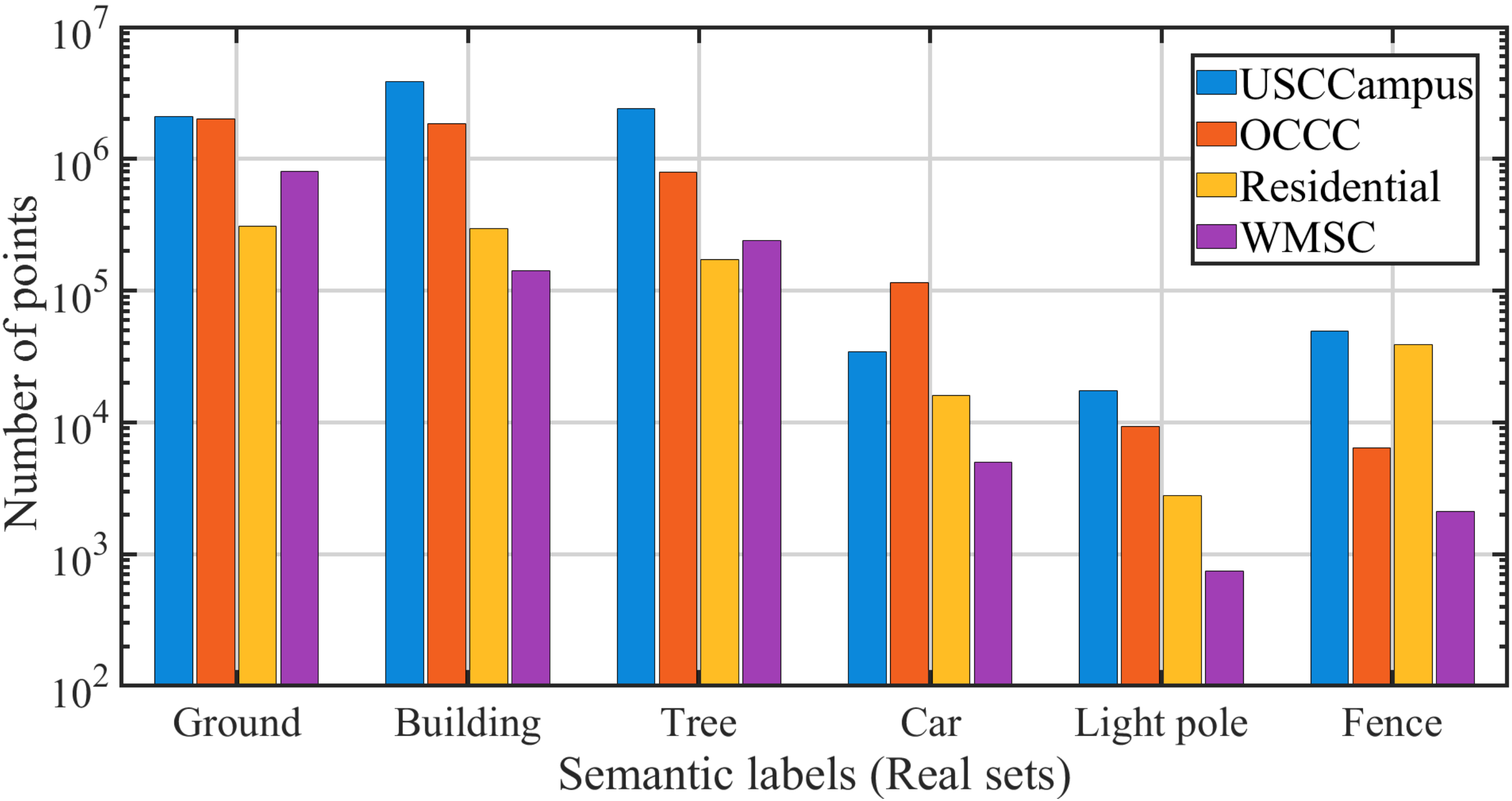}
	\end{center}
    \caption{The class distribution of \textit{real-dataset} of our \nickname{}. Note the logarithmic scale for the vertical axis.}
	\label{fig: realClass}
\end{figure}

\textbf{Real-World Datasets} To capture the real-world 3D data, we first used DJI Phantom 4 Pro for the collection of aerial images. An autonomous UAV-path planning and imagery collection system called rapid aerial photogrammetric reconstruction system (RAPTRS) \cite{60} was adopted for the survey. Specifically, RAPTRS encodes photogrammetry best practices and allows aerial photographs collected with multiple flights to cover a large area of interest. The aerial images were collected using a crosshatch-type flight pattern with predefined overlaps ranging from 75\%$\sim$85\% and flight altitudes ranging from 25m$\sim$70m. We conducted surveys on four real-world sites, including the University of Southern California Park Campus (USC), Wrigley Marine Science Center (WMSC) located on Catalina Island, Orange County Convention Center (OCCC), and a residential area (RA).

For richer diversity, we selected the area of interest to have different building and terrain types. USC campus mainly consists of commercial buildings with the paved ground (vehicle roads, pedestrian roads, and squares). Approximately 20\% of the campus is covered by grassland and tree canopy. The average height of buildings is around 5$\sim$6 floors. WMSC contains terrain with a valley and cliffs located on the shoreline of an island. OCCC is a large convention center located in the tourist district of Orlando. The surroundings of OCCC, including the parking lots and vegetated areas, are also included in our datasets. The residential area (RA) covers a typical American residential area that contains single and townhouses with an average height of 1$\sim$2 floors.

All 3D point clouds are reconstructed using the commercial software – \ieours ContextCapture. Each point is enriched with one of the six semantic class labels, including \textit{ground}, \textit{man-made structures} (including outdoor furniture, construction equipment, site storage trailers, \textit{etc.}), \textit{trees}, \textit{cars}, \textit{light poles}, and \textit{fences}. Note that the raw point clouds are subsampled to 0.3 m point spacing for training and evaluating existing segmentation methods. The statistics of the real-world dataset are also provided in Figure \ref{fig: realClass}.

\textbf{Synthetic Datasets} Following the proposed data generation pipeline, we further generated a large-scale synthetic dataset with various landscapes (\ieours various terrain shapes, types of vegetation, and urban density). It is composed of 62 point clouds and covers approximately 16 square kilometers of landscape. Additionally, the flight altitudes were set in the range of 60m$\sim$120m. Image overlaps were set in the range of 75\%$\sim$85\%, and sunlight directions were randomly assigned to simulate the data collection at different times in a day. In particular, three versions of the synthetic datasets were generated with different focuses.

\textbf{SyntheticV1}. This dataset was created using limited game objects and composed of 7 semantic categories, including \textit{ground}, \textit{building}, \textit{vegetation}, \textit{vehicle}, \textit{light pole}, \textit{street sign}, and \textit{clutter}. In this version, we mainly focused on adding diversity for large objects. For instance, details were added to the ground by randomly sculpting the input digital surface model with simulated ditches, street gutters, speed bumps, \textit{etc}. A large number of forests were also added by placing trees in random polygons with different tree spacing.

\textbf{SyntheticV2}. In this version, we focused on adding diversity for the terrain shapes and visual appearances, as well as the variation of the small objects. In particular, we selected DSM with large slopes, enriched the game objects repository, and expanded the fine-grained semantic categories. \textit{Low} (0.5m $<$height$<=$2.0m), \textit{medium} (2.0m$<$height$<=$5.0m) and \textit{high} (5.0m$<$height) vegetation class labels were adopted to separate different kinds of vegetation following ASPRS specification \cite{61}. Vehicles were further divided into passenger cars (including sedan and hatchback cars) and trucks. \textit{Bikes}, \textit{motorcycles}, \textit{fences}, \textit{roads}, \textit{aircraft}, and \textit{military vehicles} were also incorporated into the 3D scene generation process. A procedural landscape material was also leveraged to automatically generate grass and rocky textures based on the ground slope. The contextual relationship between objects was also considered, where \textit{vehicles} were placed on the \textit{roads} and \textit{light poles}, and \textit{street signs} were placed alongside the \textit{roads}. Finally, it is noted that the instance annotations for specific objects (\egours \textit{cars, trees, buildings, bikes,} \textit{etc.}) were also introduced in this version.

\begin{figure}[t]
	\begin{center}
	\includegraphics[width=1.0\linewidth]{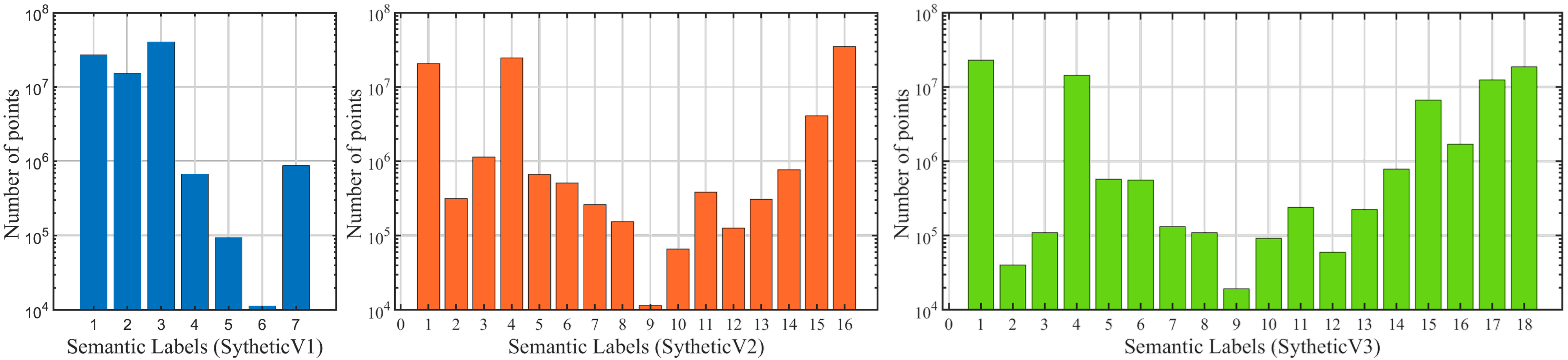}
	\end{center}
    \caption{The class distribution of \textit{synthetic} subsets of our \nickname{}. Note the logarithmic scale for the vertical axis.}
	\label{fig: syntheticClass}
\end{figure}

\begin{figure*}[tbh]
  \centering
    \includegraphics[width=0.92\linewidth]{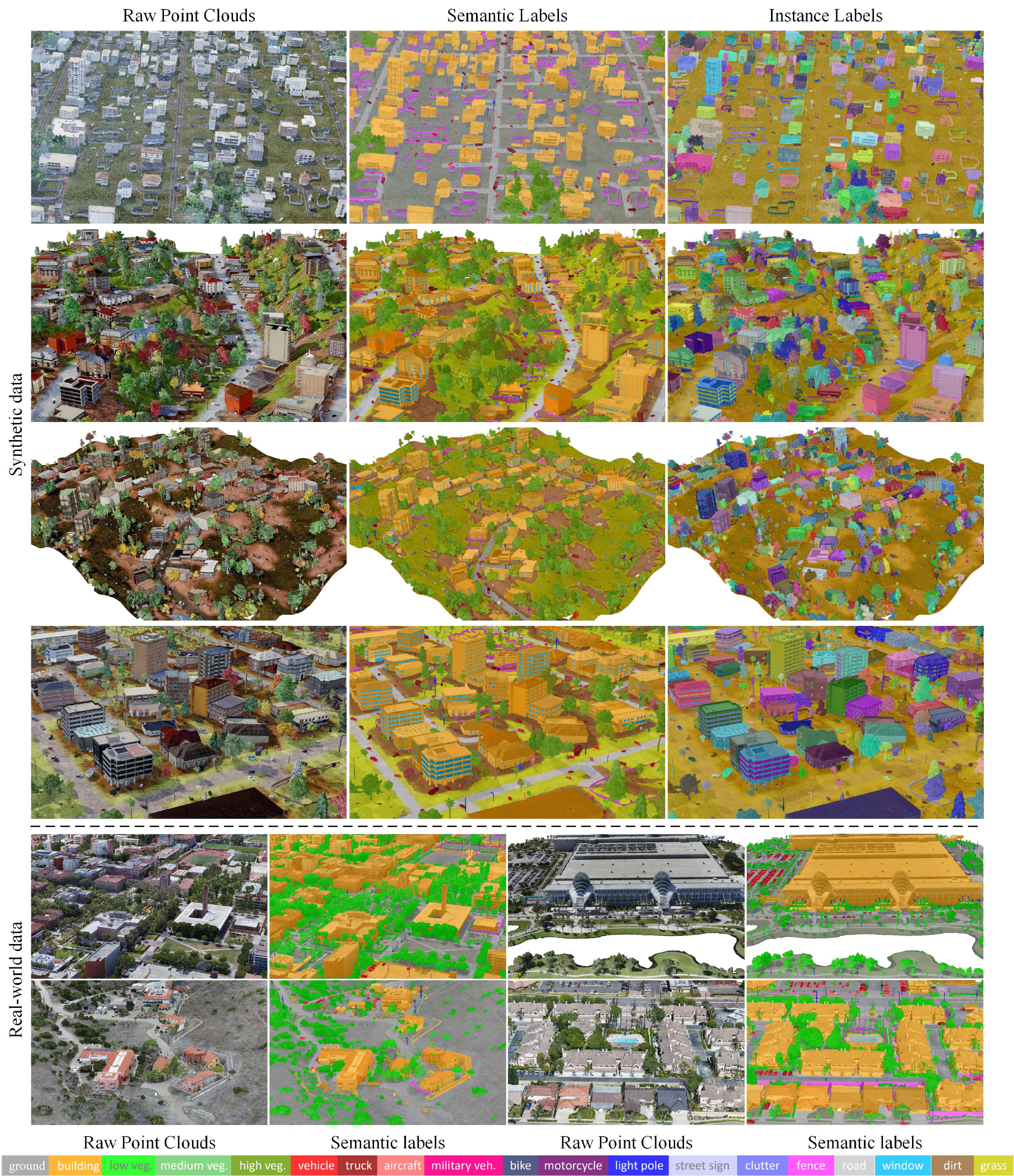}
    \caption{Examples of synthetic and real-world point clouds in our STPLS3D dataset. Different semantic classes are shown in different colors, as illustrated in the color legend. Note that different instances are displayed in different random colors. Best viewed in color.} 
    \label{figure:AdditionalVisualization}
\end{figure*}

\textbf{SyntheticV3}. In this version, we focused on large size building footprints to simulate urban areas and increase the variation of the object materials. A database of materials (including metal, rubber, signs, car paints, \textit{etc.}) for small objects (such as \textit{vehicles}, \textit{light poles}, \textit{street signs}, \textit{bikes}, \textit{motorcycles}, \textit{etc.}) was created. These materials were assigned to each object during generation. We also exploited the off-the-shelf library of photogrammetry-based textures – \textit{i.e.}, Quixel Megascans for changing materials of buildings and fences. Considering that simply assigning random materials to the facade of the building may reduce the realism of the 3D environment, we first assigned the material categories (\egours brick, concrete, wood, \textit{etc.}) to different building components (\egours wall, roof, \textit{etc.}). Next, individual material was randomly selected for each building component from the given material category. In particular, two new ground material labels (\ieours grass and dirt) were also introduced in this version. Dirt texture was painted around the building footprints with a predefined buffer, and grass texture was used to fill the blank areas that did not belong to dirt or road. The statistics of different synthetic data are shown in Figure \ref{fig: syntheticClass}. we provide additional visualization of our synthetic and real-world datasets in Figure \ref{figure:AdditionalVisualization}.

\clearpage
\section{Comparison of Data Quality}
\label{sec:ComparisonofDataQuality}
\begin{figure*}[tbh]
  \centering
    \includegraphics[width=\linewidth]{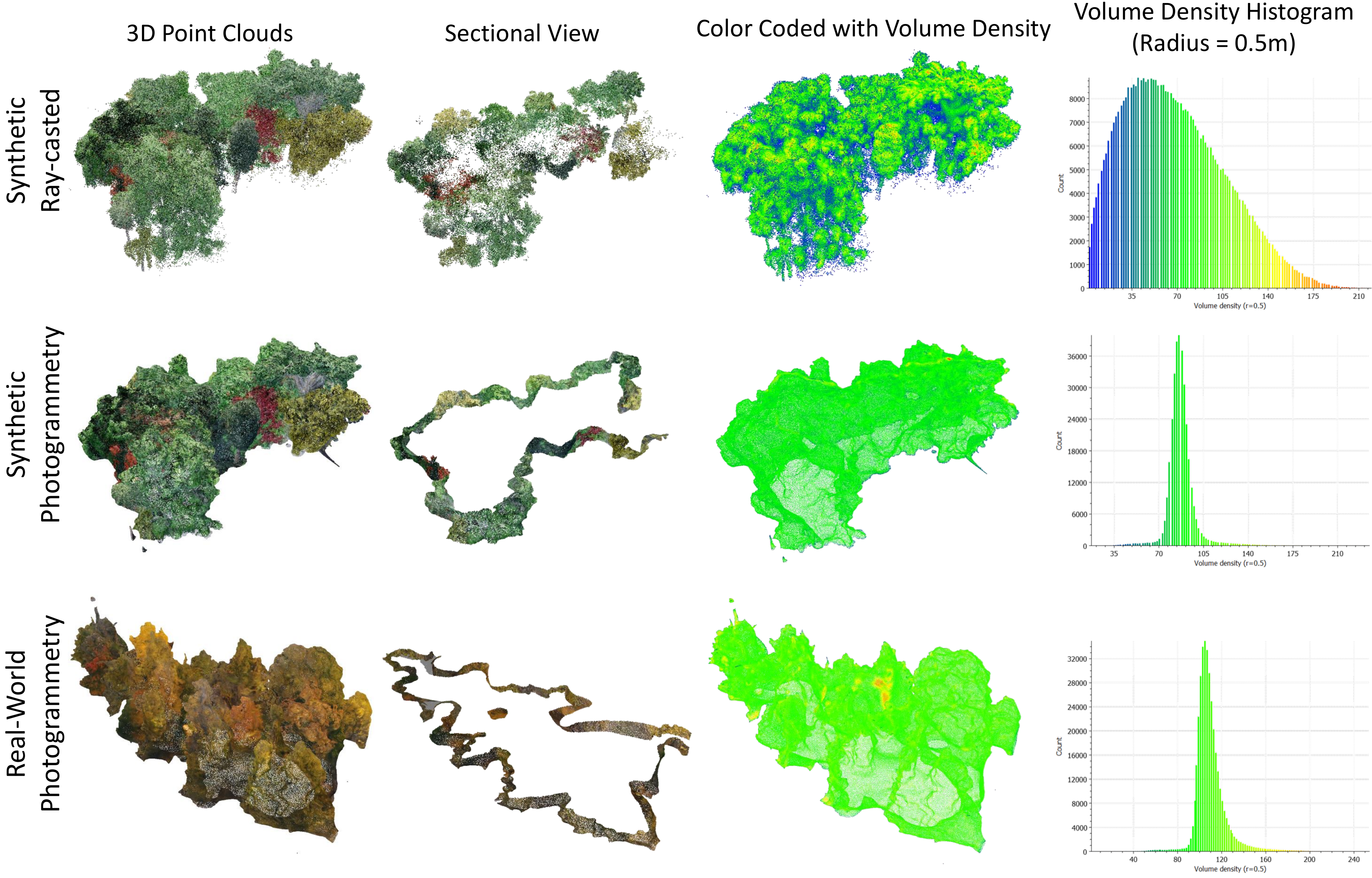}
    \caption{\qy{Qualitative comparison of tree crowns generated by ray-casted, synthetic photogrammetry, and real photogrammetry. }} 
    \label{figure:RyaVsPhotoVsReal}
\end{figure*}

\qy{As discussed in Section \ref{2D Image Rendering and 3D Reconstructions}, directly sampling or ray casting point clouds from the 3D virtual environment will lead to a large domain gap from the real photogrammetry data. Here, we provide an intuitive comparison by visualizing the ray casting 3D points, the synthetic photogrammetric points, and real-world photogrammetric points of tree crowns in Figure \ref{figure:RyaVsPhotoVsReal}. Additionally, the sectional view and volume density histogram is also reported.}

\qy{It can be seen that: 1) From the sectional view, it is clear that points of the synthetic ray-casted point clouds are scattered inside the tree crowns, while synthetic and real photogrammetry point clouds exhibit hollow-shaped shells. 2) The volume density histograms show that synthetic and real photogrammetry point clouds have much more similar point distributions compared with the synthetic ray-casted point clouds. Overall, the point clouds generated from our synthetic photogrammetry pipeline are closer to the real data. For comprehensive visualization, we also provide an anonymous video demo demonstrating the quality and distribution of different point clouds generated by ray casting, synthetical photogrammetry, and real photogrammetry. The video can be viewed at \noindent\dataQualityComparsionVideo.} 

\clearpage

\section{Object placement principles for creating the synthetic environment layouts} 
\label{Object placement principles for creating the synthetic environment layouts}

In this study, we empirically developed several simple yet effective parameterized scene layout principles. 1) Placing vehicles, trucks, bikes, and motorcycles on the paved ground and road based on the road width length with randomized intervals. 2) Placing city furniture alongside the roads with parameterized buffer sizes. 3) Placing clusters of bikes, motorcycles, and small objects around buildings and fences. 4) Scatter polygons and rings with random shapes and sizes to cover a parameterized percentage of the empty space and place objects within the same category in each polygon and ring with the parameterized minimum allowed distance. 5) Randomize the objects' rotations and scales. By mixing and matching these simple rules with parameters randomly sampled within reasonable ranges, unique strategies can be created to produce scene layouts with a large diversity and a certain realism.

\section{Evaluation of the synthetic V1, V2, and V3}
\label{sec:Evaluation of the synthetic V1, V2, and V3}

\begin{table*}[tbh]
\centering
\caption{Quantitative evaluation of different versions of the \nickname{} synthetic datasets using real subsets as testing cases. Overall Accuracy (oAcc, \%) and mean IoU (mIoU, \%) are reported.}.
\label{tab:Ablation}
\resizebox{0.9\textwidth}{!}{%
\begin{tabular}{ccccccc}
\Xhline{2.0\arrayrulewidth}
\multicolumn{1}{l}{} & \multicolumn{2}{c}{RandLA-Net \cite{57}} & \multicolumn{2}{c}{SCF-Net \cite{58}} & \multicolumn{2}{c}{KPConv \cite{46}}   \\ \Xhline{2.0\arrayrulewidth}
\multicolumn{1}{l}{} & oAcc(\%) & \textbf{mIoU(\%)} & oAcc(\%) & \textbf{mIoU(\%)} & oAcc(\%) & \textbf{mIoU(\%)}  \\ \Xhline{2.0\arrayrulewidth}
\multicolumn{1}{c|}{V1} & 78.51 & \multicolumn{1}{c|}{44.42} & 76.40 & \multicolumn{1}{c|}{43.85} & 76.71 & \multicolumn{1}{c|}{44.36}  \\
\multicolumn{1}{c|}{V2} & 80.86 & \multicolumn{1}{c|}{46.67} & 80.46 & \multicolumn{1}{c|}{46.77} & 84.18 & \multicolumn{1}{c|}{53.88}  \\
\multicolumn{1}{c|}{V3} & 76.95 & \multicolumn{1}{c|}{47.97} & 76.43 & \multicolumn{1}{c|}{47.38} & 73.42 & \multicolumn{1}{c|}{48.41}  \\ \cdashline{1-7}
\multicolumn{1}{c|}{V12} & 86.25 & \multicolumn{1}{c|}{50.41} & 86.31 & \multicolumn{1}{c|}{53.37} & 85.75 & \multicolumn{1}{c|}{51.43}  \\
\multicolumn{1}{c|}{V13} & 80.41 & \multicolumn{1}{c|}{49.08} & 79.48 & \multicolumn{1}{c|}{49.66} & 79.73 & \multicolumn{1}{c|}{51.08}  \\
\multicolumn{1}{c|}{V23} & 84.32 & \multicolumn{1}{c|}{51.07} & 85.41 & \multicolumn{1}{c|}{52.15} & 85.45 & \multicolumn{1}{c|}{55.82}  \\ \cdashline{1-7}
\multicolumn{1}{c|}{V123} & \textbf{86.39} & \multicolumn{1}{c|}{\textbf{55.76}} & \textbf{87.86} & \multicolumn{1}{c|}{\textbf{56.60}} & \textbf{88.04} & \multicolumn{1}{c|}{\textbf{58.05}} \\ \Xhline{2.0\arrayrulewidth}
\end{tabular}%
}
\end{table*}

In Section \ref{Evaluation of 3D Semantic segmentation}, we have evaluated the segmentation performance of baseline networks on our real-world dataset by training on the synthetical V1-V3 dataset. To further determine the impact of different versions of the synthetic subset on the final segmentation performance, we conducted the 7 groups of experiments as follows. Note that all groups of experiments were tested on the real-world dataset but trained with synthetic datasets with different settings.

\begin{itemize}
\setlength{\itemsep}{0pt}
\setlength{\parsep}{0pt}
\setlength{\parskip}{0pt}
    \item \qy{Trained in synthetic V1 only.}
    \item \qy{Trained in synthetic V2 only.}
    \item \qy{Trained in synthetic V3 only.}
    \item \qy{Trained in synthetic V1+V2.}
    \item \qy{Trained in synthetic V1+V3.}
    \item \qy{Trained in synthetic V2+V3.}
    \item \qy{Trained in all the synthetic V1+V2+V3.}
\end{itemize}

\qy{The quantitative performance evaluation of three baselines is shown in Table \ref{tab:Ablation}. It can be seen that: 1) The overall performance of all baselines improved when training with the combinations of synthetic subsets, compared with training on the individual synthetic subset. 2) All baselines achieved the best performance when trained on all synthetic subsets, indicating the positive impact of the synthetic datasets, especially when the real-world training data is scarce or difficult to acquire.
}

\section{Evaluation on Synthetic V3 with 18 labels}
\begin{table*}[tbh]
\centering
\setlength{\tabcolsep}{2pt}
\caption{Quantitative results on the synthetic v3 subset.}
\label{tab:syntheticv3}
\resizebox{\textwidth}{!}{%
\begin{tabular}{rccccccccccccccccccc}
\Xhline{2.0\arrayrulewidth}
 & \rotatebox{90}{\textbf{mIoU(\%)}} & \rotatebox{90}{Build.} & \rotatebox{90}{LowVeg.} & \rotatebox{90}{MediumVeg.} & \rotatebox{90}{HighVeg.} & \rotatebox{90}{Vehicle} & \rotatebox{90}{Truck} & \rotatebox{90}{Aircraft} & \rotatebox{90}{MilitaryVeh.} & \rotatebox{90}{Bike} & \rotatebox{90}{Motorcycle} & \rotatebox{90}{LightPole} & \rotatebox{90}{StreetSign} & \rotatebox{90}{Clutter} & \rotatebox{90}{Fence} & \rotatebox{90}{Road} & \rotatebox{90}{Windows} & \rotatebox{90}{Dirt} & \rotatebox{90}{Grass} \\ 
 \Xhline{2.0\arrayrulewidth}
RandLA-Net \cite{57} & 67.52 & 94.11 & 39.42 & 46.84 & 96.32 & 82.39 & 82.45 & 66.72 & 70.02 & 21.72 & 56.54 & 78.27 & 40.22 & 55.25 & 78.50 & 80.64 & 59.28 & 80.37 & \textbf{86.22} \\
SCF-Net \cite{58} & 69.07 & 93.31 & \textbf{45.40} & \textbf{49.40} & 96.51 & 82.17 & 83.08 & 68.28 & 71.93 & 22.29 & \textbf{57.98} & 80.99 & \textbf{45.27} & \textbf{63.06} & 79.29 & 79.97 & 57.63 & \textbf{80.79} & 85.86 \\
KPConv \cite{46}  & \textbf{70.35} & \textbf{96.18} & 35.17 & 47.88 & \textbf{97.04} & \textbf{86.44} & \textbf{84.24} & \textbf{75.35} & \textbf{72.26} & \textbf{23.09} & 57.59 & \textbf{86.65} & 43.07 & 62.69 & \textbf{85.60} & \textbf{81.82} & \textbf{67.23} & 79.29 & 84.68 \\ \Xhline{2.0\arrayrulewidth}
\end{tabular}%
}
\end{table*}

Here, we evaluated the segmentation performance of three baseline networks (\ieours KpConv, RandLA-Net, SCF-Net) on the synthetic subset of our dataset. In light of the data diversity and quality, we selected the SyntheticV3 dataset for evaluation. The quantitative results achieved by different baselines are shown in Table \ref{tab:syntheticv3}. It can be seen that KPConv achieved the best segmentation performance on the SyntheticV3 subset, with a mIoU score of 70.35\%, followed by the SCF-Net and RandLA-Net. We also noticed that all three baselines failed to achieve satisfactory performance on small objects such as low vegetation, bikes, and street signs. This is likely because the geometry and texture details of these small objects were lost during the 3D reconstruction in the photogrammetry process. Additionally, the performance on the underrepresented categories such as \textit{medium vegetation} is also far from satisfactory, indicating learning from imbalanced class distribution remains a challenging problem for existing techniques.

\section{Instance Segmentation with Reduced Semantic Classes}
\label{sec:Instance Segmentation with Reduced Semantic Classes}

\begin{table*}[tbh]
\setlength{\tabcolsep}{3pt}
\centering
\caption{Quantitative evaluation of two instance segmentation baselines on the synthetic v3 subset with reduced semantic classes.}
\label{tab:instance_segmentation_reduced_semantics}
\resizebox{1.0\textwidth}{!}{%
 \begin{tabular}{r|cc|ccccccccc}
 \hline
 & \multicolumn{1}{l}{Metric} & \rotatebox{90}{mean (\%)} & \rotatebox{90}{Build.} & \rotatebox{90}{Vege.} & \rotatebox{90}{Vehicle} & \rotatebox{90}{Large Vehicle} & \rotatebox{90}{Aircraft} & \rotatebox{90}{Bike} & \rotatebox{90}{Poles \& Signs} & \rotatebox{90}{Clutter} & \rotatebox{90}{Fence} \\ \hline
\multirow{3}{*}{HAIS\cite{HAIS}} & AP & 42.9 & 68.1 & 22.1 & 77.1 & 48.9 & 47.0 & 46.8 & 26.1 & 24.2 & 25.9 \\
 & AP50 & 54.6 & 74.1 & 27.2 & 87.2 & 57.8 & 67.5 & 67.0 & 34.1 & 29.5 & 47.0 \\
 & AP25 & 60.1 & 75.7 & 29.9 & 89.1 & 62.5 & 71.1 & 73.9 & 35.9 & 32.5 & 70.1 \\ \hdashline
\multirow{3}{*}{PointGroup\cite{PointGroup}} & AP & 33.4 & 63.6 & 19.8 & 57.0 & 44.4 & 36.9 & 20.0 & 21.7 & 18.2 & 19.6 \\
 & AP50 & 52.0 & 71.8 & 26.1 & 83.9 & 59.7 & 66.1 & 51.5 & 41.4 & 25.0 & 42.8 \\
 & AP25 & 61.0 & 75.4 & 30.5 & 87.0 & 64.2 & 71.6 & 70.5 & 54.7 & 28.5 & 66.7 \\ \hline
\end{tabular}%
}
\end{table*}

Since the instance segmentation performance depends on the quality of the semantic segmentation results, We also provided an instance segmentation benchmark with reduced semantics by merging similar semantic classes to eliminate the cascade effect from poor semantic segmentation. Specifically, \textit{low}, \textit{medium}, and \textit{high vegetation} were merged into the \textit{vegetation} category. \textit{Bicycle} and \textit{motorcycle} points were labeled as \textit{bike}. \textit{Trucks} and \textit{military vehicles} were combined as \textit{large vehicles}. The \textit{street signs} were joined with \textit{light poles}. Thus, the semantics were reduced from 14 classes to 9 classes. PointGroup \cite{PointGroup} and HAIS \cite{HAIS} were used again as the baselines. The results are shown in Table \ref{tab:instance_segmentation_reduced_semantics}. Both HAIS and PointGroup achieved a higher AP (nearly 9\%), which shows that if the semantic segmentation capability for similar object classes could be improved in the end-to-end instance segmentation networks, the performance of instance segmentation could also be increased.

\section{Data-Related Hyperparameters of Benchmark Methods}
\label{Data-Related Hyperparameters of Benchmark Methods}
To achieve a trade-off between data scale, resolution, and computing resources, we empirically set 0.3m for grid downsampling to reduce the number of total points while preserving enough details. In addition, for voxel-based approaches, including MinkowskiNet, PointGroup, and HAIS, we set the sample size of $\SI{50}{\metre} \times \SI{50}{\metre}$ on the XY plane. We used a sample size of $\SI{100}{\metre} \times \SI{100}{\metre}$ for PointTransformer, an 18m radius of sphere for KpConv, and 40,960 input points for SCF-Net and RandLA-Net.


\section{Definition of Semantic Categories} 
\label{sec:Definition of Semantic Categories}

\qy{Here, we provide a detailed definition of the semantic categories in our \nickname{} dataset.}
\newline
\newline
\indent\textbf{SyntheticV1:}
\begin{enumerate}[leftmargin=1cm]
\setlength{\itemsep}{0pt}
\setlength{\parsep}{0pt}
\setlength{\parskip}{0pt}
  \item Ground: including grass, paved road, dirt, etc.
  \item Building: including commercial, residential, educational buildings.
  \item Vegetation: including low, medium, and high vegetation.
  \item Vehicle: including sedan and hatchback cars.
  \item Light pole: including light poles and traffic lights.
  \item Street sign: including road signs at the side of roads.
  \item Clutter: including city furniture, construction equipment, barricades, and other 3D shapes.
\end{enumerate}

\textbf{SyntheticV2}:
\begin{enumerate}[leftmargin=1cm]
\setlength{\itemsep}{0pt}
\setlength{\parsep}{0pt}
\setlength{\parskip}{0pt}
  \item Building: Same as the definition of building in SyntheticV1.
  \item Low vegetation: 0.5 m $<$ vegetation height $<$ 2.0 m.
  \item Medium vegetation: 2.0 m $<$ vegetation height $<$ 5.0 m.
  \item High vegetation: 5.0 m $<$ vegetation height.
  \item Passenger car: including sedans and hatchback cars.
  \item Truck: including pickup trucks, cement trucks, flat-bed trailers, trailer trucks, etc.
  \item Aircraft: including helicopters and airplanes
  \item Military vehicle: including tanks and Humvees.
  \item Bike: bicycles.
  \item Motorcycle: motorcycles.
  \item Light pole: Same as the definition of light pole in SyntheticV1.
  \item Street sign: Same as the definition of street sign in SyntheticV1.
  \item Clutter: Same as the definition of clutter in SyntheticV1.
  \item Fence: including timber, brick, concrete, metal fences.
  \item Road: including asphalt and concrete roads.
  \item Grass: including grass lawn, wild grass, etc.
\end{enumerate}

\textbf{SyntheticV3}:
\begin{enumerate}[leftmargin=1cm]
\setlength{\itemsep}{0pt}
\setlength{\parsep}{0pt}
\setlength{\parskip}{0pt}
  \setcounter{enumi}{15}
  \item Window: glass windows.
  \item Dirt: bare earth.
  \item Grass: Same as the definition of grass in SyntheticV2.
\end{enumerate}

Note that, the definition of classes 1 to 15 is the same as SyntheticV2.
\newline
\newline
\indent\textbf{Real-world data}:
\begin{enumerate}[leftmargin=1cm]
\setlength{\itemsep}{0pt}
\setlength{\parsep}{0pt}
\setlength{\parskip}{0pt}
  \item Ground: including grass, paved roads, dirt, sidewalk, parking lots, etc.
  \item Tree: including low, medium, and high vegetation.
  \item Car: including sedans and hatchback cars, pickup trucks, flatbed trailers, trailer trucks, etc.
  \item Light pole: including light poles, traffic lights, and street signs.
  \item Fence: including timber, brick, concrete, metal fences.
  \item Building (man-made structure): Including buildings, city furniture,
construction equipment, site storage trailers, \textit{etc.} (\ieours Objects that do not belong to ground, tree, car, light pole, and fence.)
\end{enumerate}

\section{Class mapping between synthetic and real data}
\label{Class mapping between synthetic and real data}
Considering that the semantic categories of the synthetic datasets are inconsistent with the real-world dataset (18 vs. 6), we conducted a class mapping to unify the semantic categories for the experiments discussed in \ref{Evaluation of 3D Semantic segmentation}. Specifically, \textit{road}, \textit{dirt}, and \textit{grass} points were combined as \textit{ground}. \textit{Low}, \textit{medium} and \textit{high vegetation} were merged into the \textit{vegetation} category. \textit{Cars}, \textit{trucks}, and \textit{military vehicles} were labeled as \textit{vehicles}. The \textit{street sign} was joined with \textit{light poles}, and all other objects except fences were merged with buildings as \textit{man-made structures}.

\section{Visualization of the FDc dataset} 
\label{Visualization of the FDc dataset}
\begin{figure*}[t]
  \centering
    \includegraphics[width=\linewidth]{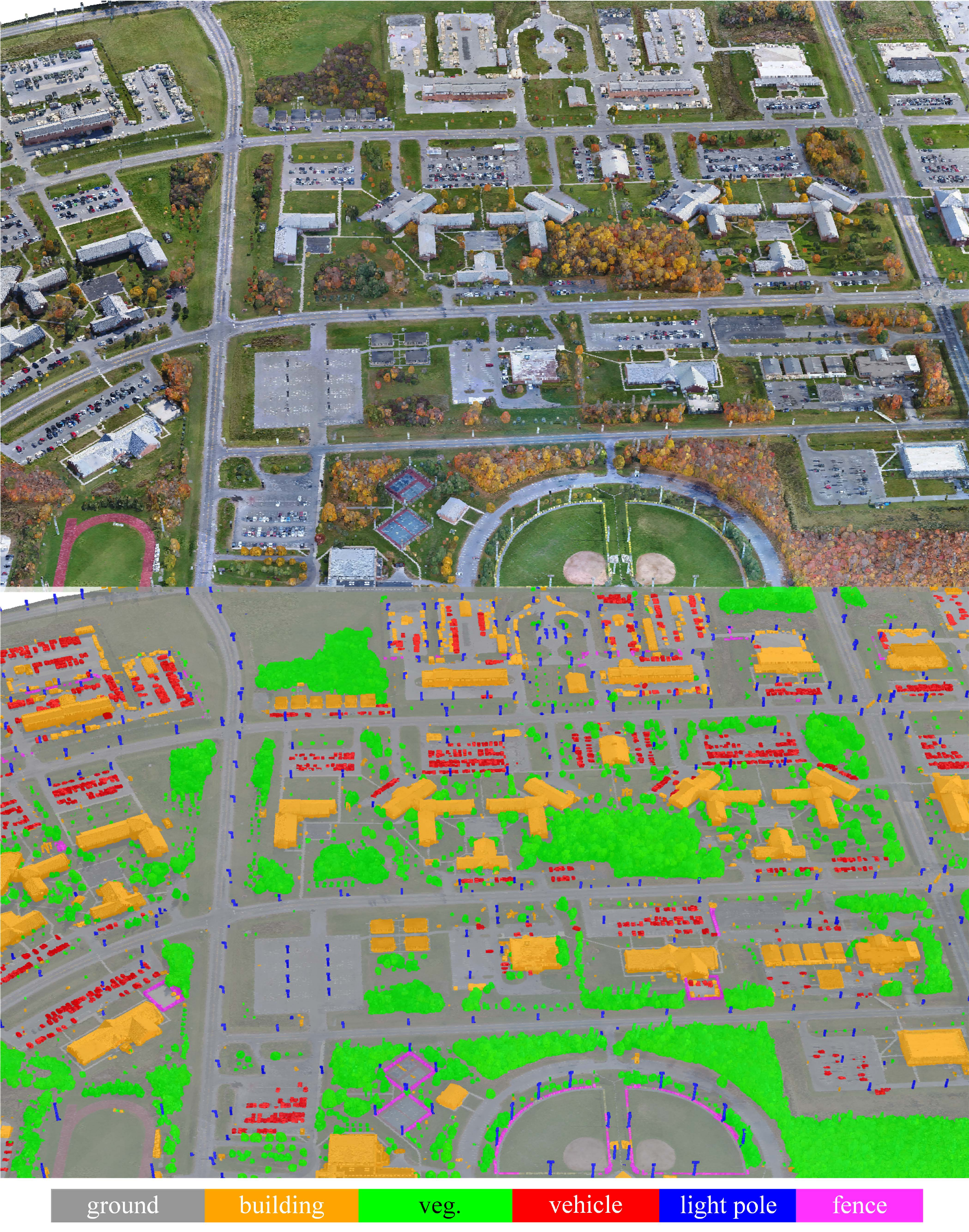}
    \caption{Example visualization of the FDc dataset.}
    \label{figure:FDc}
\end{figure*}

To have an intuitive and clear understanding of the FDc data, we visualize the 3D point cloud along with its annotations in Figure \ref{figure:FDc}.

\end{document}